\documentclass[10pt,twocolumn,letterpaper]{article}

\usepackage[pagenumbers]{iccv} %

\usepackage[ruled, lined, linesnumbered, commentsnumbered, longend]{algorithm2e}
\usepackage{multirow}

\newcommand{\OURS}{3DGS-LM}
\newcommand{\mypar}[1]{\vspace{0.5mm}\noindent\textbf{#1}}
\newcommand{\scene}[1]{\texttt{\small{#1}}}

\definecolor{iccvblue}{rgb}{0.21,0.49,0.74}
\usepackage[pagebackref,breaklinks,colorlinks,allcolors=iccvblue]{hyperref}
\usepackage[accsupp]{axessibility}  %

\title{\OURS: Faster Gaussian-Splatting Optimization with Levenberg-Marquardt}

\author{
Lukas H{\"o}llein$^{1}$ \quad
Alja\v{z} Bo\v{z}i\v{c}$^{2}$ \quad
Michael Zollh{\"o}fer$^2$ \quad
Matthias Nie{\ss}ner$^1$ \\[0.5em]
$^1$Technical University of Munich \quad $^2$Meta \\
\url{https://lukashoel.github.io/3DGS-LM/} \\[-0.8em]
}
\begin{document}
\begin{figure}
\vspace{-4mm}
\twocolumn[{
\renewcommand\twocolumn[1][]{#1}
\maketitle

\centering
\setlength\tabcolsep{0pt}
\begin{tabular}{c@{}c}
\includegraphics[width=0.4319\textwidth]{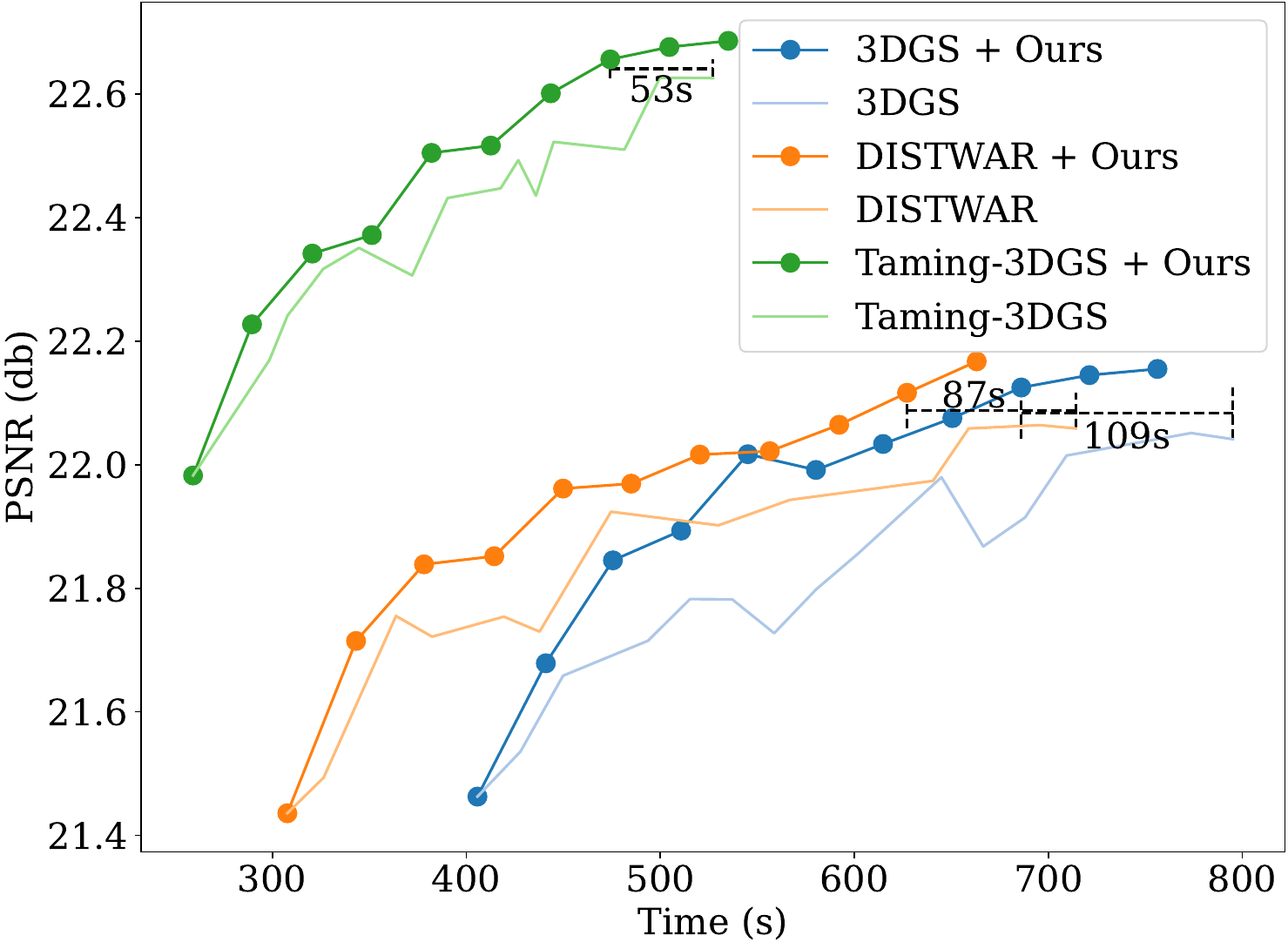} &
\includegraphics[width=0.5681\textwidth]{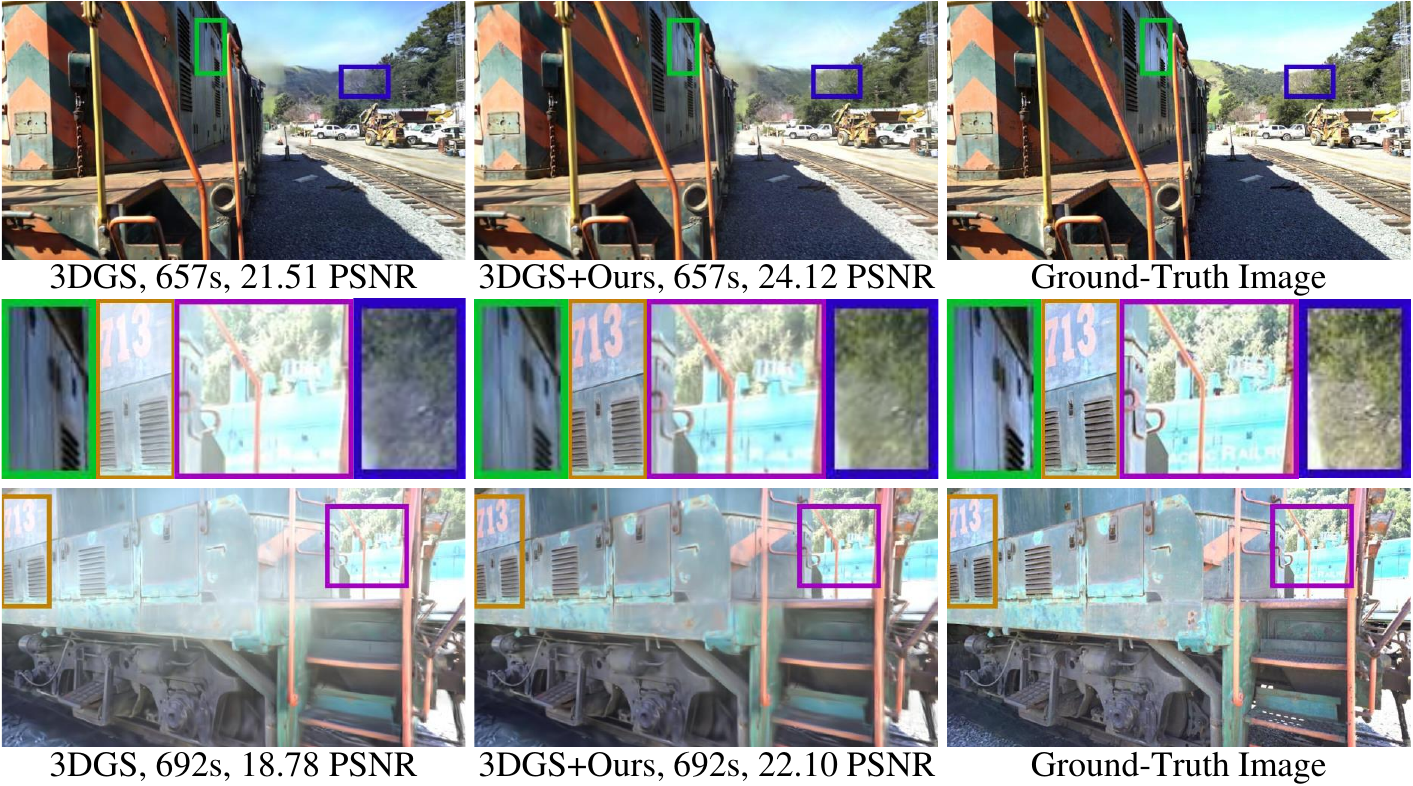} \\
\end{tabular}
\vspace{-1.2mm}
\caption{
Our method accelerates 3D Gaussian Splatting (3DGS)~\cite{kerbl3Dgaussians} reconstruction by replacing the ADAM optimizer with a tailored Levenberg-Marquardt.
Left: starting from the same initialization, our method converges faster on the Tanks\&Temples \scene{TRAIN} scene.
Right: after the same amount of time, our method produces higher quality renderings (e.g., better brightness and contrast).
}
\vspace{2mm}
\label{fig:teaser}
}]
\end{figure}

\begin{abstract}
We present \OURS, a new method that accelerates the reconstruction of 3D Gaussian Splatting (3DGS) by replacing its ADAM optimizer with a tailored Levenberg-Marquardt (LM).
Existing methods reduce the optimization time by decreasing the number of Gaussians or by improving the implementation of the differentiable rasterizer.
However, they still rely on the ADAM optimizer to fit Gaussian parameters of a scene in thousands of iterations, which can take up to an hour.
To this end, we change the optimizer to LM that runs in conjunction with the 3DGS differentiable rasterizer. %
For efficient GPU parallelization, we propose a caching data structure for intermediate gradients that allows us to efficiently calculate Jacobian-vector products in custom CUDA kernels.
In every LM iteration, we calculate update directions from multiple image subsets using these kernels and combine them in a weighted mean.
Overall, our method is 20\% faster than the original 3DGS while obtaining the same reconstruction quality.
Our optimization is also agnostic to other methods that accelerate 3DGS, thus enabling even faster speedups compared to vanilla 3DGS.
\end{abstract}

\section{Introduction}
\label{sec:intro}
Novel View Synthesis (NVS) is the task of rendering a scene from new viewpoints, given a set of images as input.
NVS can be employed in Virtual Reality applications to achieve photo-realistic immersion and to freely explore captured scenes.
To facilitate this, different 3D scene representations have been developed~\cite{mildenhall2021nerf, kerbl3Dgaussians, muller2022instant, barron2021mip, Tewari2022NeuRendSTAR, aliev2020neural}.
Among those, 3DGS~\cite{kerbl3Dgaussians} (3D Gaussian-Splatting) is a point-based representation that parameterizes the scene as a set of 3D Gaussians.
It offers real-time rendering and high-quality image synthesis, while being optimized from a set of posed images through a differentiable rasterizer.

3DGS is optimized from a set of posed input images that densely capture the scene.
The optimization can take up to an hour to converge on high-resolution real-world scene datasets with a lot of images~\cite{yeshwanth2023scannet++}.
It is desirable to reduce the optimization runtime which enables faster usage of the reconstruction for downstream applications. 
Existing methods reduce this runtime by improving the optimization along different axes.
First, methods accelerate the rendering speed of the tile-based, differentiable rasterizer or the backward-pass that is specifically tailored for optimization with gradient descent~\cite{durvasula2023distwar, mallick2024taming, ye2024gsplatopensourcelibrarygaussian, feng2024flashgs}.
For example, Durvasula~\etal~\cite{durvasula2023distwar} employ warp reductions for a more efficient sum of rendering gradients, while Mallick~\etal~\cite{mallick2024taming} utilizes a splat-parallelization for backpropagation.
Second, in 3DGS the number of Gaussians is gradually grown during optimization, which is known as densification.
Recently, GS-MCMC~\cite{kheradmand20243d}, Taming-3DGS~\cite{mallick2024taming}, Mini-Splatting~\cite{fang2024mini}, and Revising-3DGS~\cite{bulo2024revising} propose novel densification schemes that reduce the number of required Gaussians to represent the scene.
This makes the optimization more stable and also faster, since fewer Gaussians must be optimized and rendered in every iteration.

Despite these improvements, the optimization still takes significant resources, requiring thousands of gradient descent iterations to converge.
To this end, we aim to reduce the runtime by improving the underlying optimization during 3DGS reconstruction.
More specifically, we propose to replace the widely used ADAM~\cite{kingma2014adam} optimizer with a tailored Levenberg-Marquardt (LM)~\cite{more2006levenberg}.
LM is known to drastically reduce the number of iterations by approximating second-order updates through solving the normal equations (\cref{tab:abl-mv-constraints}).
This allows us to accelerate 3DGS reconstruction (\cref{fig:teaser} left) by over 20\% on average.
Concretely, we propose a highly efficient GPU parallelization scheme for the preconditioned conjugate gradient (PCG) algorithm within the inner LM loop in order to obtain the respective update directions. %
To this end, we extend the differentiable 3DGS rasterizer with custom CUDA kernels that compute Jacobian-vector products.
Our proposed caching data structure for intermediate gradients (\cref{fig:parallelization}) then allows us to perform these calculations fast and efficiently in a data-parallel fashion.
In order to scale caching to high-resolution image datasets, we calculate update directions from multiple image subsets and combine them in a weighted mean.
Overall, this allows us to improve reconstruction time by 20\% compared to state-of-the-art 3DGS baselines while achieving the same reconstruction quality (\cref{fig:teaser} right).

\medskip
\noindent To summarize, our contributions are:
\begin{itemize}
    \item we propose a tailored 3DGS optimization based on Levenberg-Marquardt that improves reconstruction time by 20\% and which is agnostic to other 3DGS acceleration methods.
    \item we propose a highly-efficient GPU parallelization scheme for the PCG algorithm for 3DGS in custom CUDA kernels with a caching data structure to facilitate efficient Jacobian-vector products. %
\end{itemize}

\section{Related Work}
\label{sec:related-work}

\subsection{Novel-View-Synthesis}
Novel-View-Synthesis is widely explored in recent years \cite{hedman2018deep, aliev2020neural, Tewari2022NeuRendSTAR, mildenhall2021nerf, kerbl3Dgaussians, barron2021mip, muller2022instant}.
NeRF~\cite{mildenhall2021nerf} achieves highly photo-realistic image synthesis results through differentiable volumetric rendering.
It was combined with explicit representations to accelerate optimization runtime \cite{Fridovich-Keil_2022_CVPR, sun2022direct, muller2022instant, xu2022point, chen2022tensorf}.
3D Gaussian Splatting (3DGS)~\cite{kerbl3Dgaussians} extends this idea by representing the scene as a set of 3D Gaussians, that are rasterized into 2D splats and then $\alpha$-blended into pixel colors.
The approach gained popularity, due to the ability to render high quality images in real-time.
Since its inception, 3DGS was improved along several axes.
Recent methods improve the image quality by increasing or regularizing the capacity of primitives~\cite{Yu2024MipSplatting, hamdi2024ges, scaffoldgs, held20243d, huang2024deformable}.
Others increase rendering efficiency~\cite{niemeyer2024radsplat, ren2024octree}, obtain better surface reconstructions~\cite{Huang2DGS2024, guedon2023sugar}, reduce the memory requirements~\cite{papantonakis2024reducing}, and enable large-scale reconstruction~\cite{hierarchicalgaussians24, zhao2024scaling3dgaussiansplatting}.
We similarly adopt 3DGS as our scene representation and focus on improving the per-scene optimization runtime.

\subsection{Speed-Up Gaussian Splatting Optimization}
Obtaining a 3DGS scene reconstruction can be accelerated in several ways.
One line of work reduces the number of Gaussians by changing the densification heuristics \cite{kheradmand20243d, mallick2024taming, fang2024mini, bulo2024revising, scaffoldgs, lu2024turbo}.
Other methods focus on sparse-view reconstruction and train a neural network as data prior, that outputs Gaussians in a single forward pass \cite{fan2024instantsplat, liu2024mvsgaussian, chen2024mvsplat, LaRa, charatan2024pixelsplat, xu2024depthsplat, ziwen2024long}.
In contrast, we focus on the dense-view and per-scene optimization setting, i.e., we are not limited to sparse-view reconstruction.
Most related are methods that improve the implementation of the underlying differentiable rasterizer.
In \cite{durvasula2023distwar, ye2024gsplatopensourcelibrarygaussian} the gradient descent backward pass is accelerated through warp-reductions, while \cite{mallick2024taming} improves its parallelization pattern and \cite{feng2024flashgs} accelerates the rendering.
In contrast, we completely replace the gradient descent optimization with LM through a novel and tailored GPU parallelization scheme.
We demonstrate that we are compatible with those existing methods, i.e., we further reduce runtime by plugging our optimizer into their scene initializations.

\subsection{Optimizers For 3D Reconstruction Tasks}
NeRF and 3DGS are typically optimized with stochastic gradient descent (SGD) optimizers like ADAM~\cite{kingma2014adam} for thousands of iterations.
In contrast, many works in RGB-D fusion employ the Gauss-Newton (or Levenberg-Marquardt) algorithms to optimize objectives for 3D reconstruction tasks~\cite{zollhofer2014real, zollhofer2015shading, Thies_2016_CVPR, thies2015real, dai2017bundlefusion, devito2017opt}.
By doing so, these methods can quickly converge in an order of magnitude fewer iterations than SGD.
Motivated by this, we aim to accelerate 3DGS optimization by adopting the Levenberg-Marquardt algorithm as our optimizer.
Rasmuson \etal~\cite{rasmuson2022perf} implemented the Gauss-Newton algorithm for reconstructing low-resolution NeRFs based on dense voxel grids.
In contrast, we exploit the explicit Gaussian primitives of 3DGS to perform highly-efficient Jacobian-vector products in a data-parallel fashion.
This allows us to achieve state-of-the-art rendering quality, while significantly accelerating the optimization in comparison to ADAM-based methods.

\section{Method}
\label{sec:method}
\begin{figure}
\includegraphics[width=0.5\textwidth]{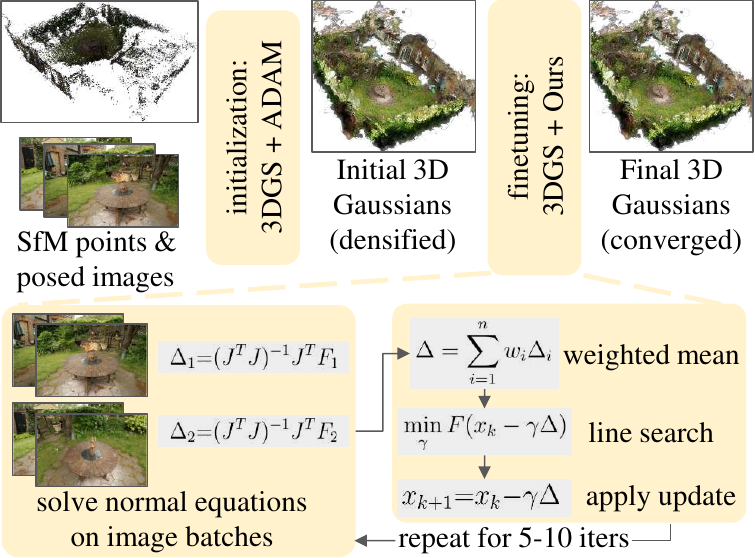}
\caption{
\textbf{Method Overview.}
We accelerate 3DGS optimization by framing it in two stages.
First, we use the original ADAM optimizer and densification scheme to arrive at an initialization for all Gaussians.
Second, we employ the Levenberg-Marquardt algorithm to finish optimization.
}
\label{fig:pipeline}
\end{figure}

Our pipeline is visualized in ~\cref{fig:pipeline}.
First, we obtain an initialization of the Gaussians from a set of posed images and their SfM point cloud as input by running the standard 3DGS optimization (\cref{subsec:3dgs}).
In this stage the Gaussians are densified, but remain unconverged.
Afterwards, we finish the optimization with our novel optimizer.
Concretely, we optimize the sum of squares objective with the Levenberg-Marquardt (LM)~\cite{more2006levenberg} algorithm (\cref{subsec:gsgn-theory}), which we implement in efficient CUDA kernels (\cref{subsec:gsgn-kernels}).
This two-stage approach accelerates the optimization compared to only using first-order optimizers.

\subsection{Review Of Gaussian-Splatting}
\label{subsec:3dgs}
3D Gaussian Splatting (3DGS) ~\cite{kerbl3Dgaussians} models a scene as a set of 3D Gaussians, each of which is parameterized by a position, rotation, scaling, and opacity. 
The view-dependent color is modeled by Spherical Harmonics coefficients of order 3.
To render an image of the scene from a given viewpoint, all Gaussians are first projected into 2D Gaussian splats with a tile-based differentiable rasterizer.
Afterwards, they are $\alpha$-blended along a ray to obtain the pixel color $c$:
\begin{align}
\label{eq:alpha-blending}
 c = \sum_{i \in \mathcal{N}} c_i \alpha_i T_i, \quad \text{with  } T_i = \prod_{j=1}^{i-1} (1 - \alpha_j)
\end{align}
where $c_i$ is the color of the $i$-th splat along the ray, $\alpha_i$ is given by evaluating the 2D Gaussian multiplied with its opacity, and $T_i$ is the transmittance.
To fit all Gaussian parameters $\mathbf{x} \in \mathbb{R}^M$ to posed image observations, a rendering loss is minimized with the ADAM~\cite{kingma2014adam} optimizer:
\begin{align}
\label{eq:sgd-loss}
\mathcal{L}(\mathbf{x}) {=} \frac{1}{N} \sum_{i=1}^{N} (\lambda_{1} |c_i {-} C_i| {+} \lambda_{2} (1 {-} \text{SSIM}(c_i, C_i)))
\end{align}
where $\lambda_1 {=} 0.8$, $\lambda_2 {=} 0.2$, and $C_i$ the ground-truth for one pixel.
Typically, 3DGS uses a batch size of 1 by sampling a random image per update step.
The Gaussians are initialized from the SfM points and their number is gradually grown during the first half of the optimization, which is known as densification~\cite{kerbl3Dgaussians}.

\subsection{Levenberg-Marquardt Optimization For 3DGS}
\label{subsec:gsgn-theory}
We employ the LM algorithm for optimization of the Gaussians by reformulating the rendering loss as a sum of squares energy function:
\begin{align}
\label{eq:gn-loss}
E(\mathbf{x}) {=} \sum_{i=1}^{N} \sqrt{\lambda_1 |c_i {-} C_i|}^2 {+} \sqrt{\lambda_2 (1 {-} \text{SSIM}(c_i, C_i))}^2
\end{align}
where we have two \textit{separate} residuals $r^{\text{abs}}_i {=} \sqrt{\lambda_1 |c_i {-} C_i|}$ and $r^{\text{SSIM}}_i {=} \sqrt{\lambda_2 (1 {-} \text{SSIM}(c_i, C_i))}$ per color channel of each pixel.
We take the square root of each loss term, to convert \cref{eq:sgd-loss} into the required form for the LM algorithm.
In other words, we use the identical objective, but a different optimizer.
In contrast to ADAM, the LM algorithm requires a large batch size (ideally all images) for every update step to achieve stable convergence~\cite{more2006levenberg}.
In practice, we select large enough subsets of all images to ensure reliable update steps (see \cref{subsec:gsgn-kernels} for more details).

\mypar{Obtaining Update Directions}
In every iteration of our optimization we obtain the update direction $\Delta \in \mathbb{R}^M$ for all $M$ Gaussian parameters by solving the normal equations:
\begin{align}
\label{eq:norm-eq}
(\mathbf{J}^T \mathbf{J} + \lambda_\text{reg} \text{diag}(\mathbf{J}^T \mathbf{J})) \Delta = -\mathbf{J}^T \mathbf{F}(\mathbf{x})
\end{align}
where $\mathbf{F}(\mathbf{x}) {=} [r^{\text{abs}}_1, ..., r^{\text{abs}}_N, r^{\text{SSIM}}_1, ..., r^{\text{SSIM}}_N] {\in} \mathbb{R}^{2N}$ is the residual vector corresponding to \cref{eq:gn-loss} and $\mathbf{J} {\in} \mathbb{R}^{2N {\times} M}$ the corresponding Jacobian matrix.

In a typical dense capture setup, we optimize over millions of Gaussians and have hundreds of high-resolution images~\cite{knapitsch2017tanks, hedman2018deep, barron2022mip}.
Even though $\mathbf{J}$ is a sparse matrix (each row only contains non-zero values for the Gaussians that contribute to the color of that pixel), it is therefore not possible to materialize $\mathbf{J}$ in memory.
Instead, we employ the preconditioned conjugate gradient (PCG) algorithm, to solve \cref{eq:norm-eq} in a \textit{matrix-free} fashion.
We implement PCG in custom CUDA kernels, see \cref{subsec:gsgn-kernels} for more details.

\mypar{Apply Parameter Update}
After we obtained the solution $\Delta$, we run a line search to find the best scaling factor $\gamma {\in} \mathbb{R}$ for updating the Gaussian parameters:
\begin{align}
\label{eq:line-search}
\min_\gamma E(\mathbf{x}_k + \gamma \Delta)
\end{align}
In practice, we run the line search on a 30\% subset of all images, which is enough to get a reasonable estimate for $\gamma$, but requires fewer rendering passes.
Afterwards, we update the Gaussian parameters as: $\mathbf{x}_{k+1} {=} \mathbf{x}_k + \gamma \Delta$.
Similar to the implementation of LM in CERES~\cite{Agarwal_Ceres_Solver_2022}, we adjust the regularization strength $\lambda_\text{reg} {\in} \mathbb{R}$ after every iteration based on the quality of the update step.
Concretely, we calculate
\begin{align}
\label{eq:tr-update}
\rho = \frac{||\mathbf{F}(\mathbf{x})||^2 - ||\mathbf{F}(\mathbf{x} + \gamma \Delta)||^2}{||\mathbf{F}(\mathbf{x})||^2 - ||\mathbf{J} \gamma \Delta + \mathbf{F}(\mathbf{x})||^2}
\end{align}
and only keep the update if $\rho {>} 1\mathrm{e}{-5}$, in which case we reduce the regularization strength as $\lambda_\text{reg} {\mathrel{*}=} 1 {-} (2 \rho -1)^3$.
Otherwise, we revert the update and double $\lambda_\text{reg}$.

\subsection{Efficient Parallelization Scheme For PCG}
\label{subsec:gsgn-kernels}

The PCG algorithm obtains the solution to the least squares problem of~\cref{eq:norm-eq} in multiple iterations.
We run the algorithm for up to $\text{n}_\text{iters} {=} 8$ iterations and implement it with custom CUDA kernels.
We summarize it in~\cref{alg:pcg}.

\begin{algorithm}
    \SetKwFunction{buildCache}{\textcolor{blue}{buildCache}}
    \SetKwFunction{sortCacheBySplats}{\textcolor{blue}{sortCacheByGaussians}}
    \SetKwFunction{diagJTJ}{\textcolor{blue}{diagJTJ}}
    \SetKwFunction{sortX}{\textcolor{blue}{sortX}}
    \SetKwFunction{applyJ}{\textcolor{blue}{applyJ}}    
    \SetKwFunction{applyJT}{\textcolor{blue}{applyJT}}
    
    \SetKwInOut{KwIn}{Input}
    \SetKwInOut{KwOut}{Output}
    \KwIn{Gaussians and cameras $\mathcal{G}$, $\mathbf{F}$, $\lambda_\text{reg}$}
    \KwOut{Update direction $\Delta$}

    $\mathbf{b}, \mathcal{C} = \buildCache(\mathcal{G}, \mathbf{F})$  \tcp*[f]{$\mathbf{b} {=} -\mathbf{J}^T \mathbf{F}$}

    $\mathcal{C} = \sortCacheBySplats(\mathcal{C})$

    $\mathbf{M}^{-1} = 1 / \diagJTJ(\mathcal{G}, \mathcal{C})$

    $\mathbf{x_0} = \mathbf{M}^{-1} \mathbf{b}$

    $\mathbf{u}_0 = \applyJ(\sortX(\mathbf{x}_0), \mathcal{G}, \mathcal{C})$  \tcp*[f]{$\mathbf{u}_0 {=} \mathbf{J} \mathbf{x}_0$}
    
    $\mathbf{g}_0 = \applyJT(\mathbf{u}_0, \mathcal{G}, \mathcal{C})$  \tcp*[f]{$\mathbf{g}_0 {=} \mathbf{J}^T \mathbf{u}_0$}

    $\mathbf{r}_0 = \mathbf{b} - (\mathbf{g}_0 {+} \lambda_\text{reg} \mathbf{M} \mathbf{x}_0)$

    $\mathbf{z}_0 = \mathbf{M}^{-1} \mathbf{r}_0$
    
    $\mathbf{p}_0 = \mathbf{z}_0$

    \For{$i = 0$ \KwTo $\text{n}_\text{iters}$}{

        $\mathbf{u}_i = \applyJ(\sortX(\mathbf{p}_i), \mathcal{G}, \mathcal{C})$  \tcp*[f]{$\mathbf{u}_i {=} \mathbf{J} \mathbf{p}_i$}
        
        $\mathbf{g}_i = \applyJT(\mathbf{u}_i, \mathcal{G}, \mathcal{C})$  \tcp*[f]{$\mathbf{g}_i {=} \mathbf{J}^T \mathbf{u}_i$}

        $\mathbf{g}_i \mathrel{+}= \lambda_\text{reg} \mathbf{M} \mathbf{p}_i$
        
        $\alpha_{i} = \frac{\mathbf{r}_i^T \mathbf{z}_i}{\mathbf{p}_i^T \mathbf{g}_i}$

        $\mathbf{x}_{i+1} {=} \mathbf{x}_i {+} \alpha_i \mathbf{p}_i$
        
        $\mathbf{r}_{i+1} {=} \mathbf{r}_i {-} \alpha_i \mathbf{g}_i$
        
        $\mathbf{z}_{i+1} {=} \mathbf{M}^{-1} \mathbf{r}_{i+1}$

        $\beta_i = \frac{\mathbf{r}^T_{i+1} \mathbf{z}_{i+1}}{\mathbf{r}^T_{i} \mathbf{z}_{i}}$

        $\mathbf{p}_{i+1} = \mathbf{z}_{i+1} + \beta_i \mathbf{p}_i$

        \If{$||\mathbf{r}_{i+1}||^2 < 0.01 ||\mathbf{b}||^2$}{break}
    }

    \KwRet{$\mathbf{x}_{i+1}$}
    \caption{We run the PCG algorithm with custom CUDA kernels (blue) in every LM iteration.}
    \label{alg:pcg}
\end{algorithm}

Most of the work in every PCG iteration is consumed by calculating the matrix-vector product $\mathbf{g}_i {=} \mathbf{J}^T \mathbf{J} \mathbf{p}_i$.
We compute it by first calculating $\mathbf{u}_i {=} \mathbf{J} \mathbf{p}_i$ and then $\mathbf{g}_i {=} \mathbf{J}^T \mathbf{u}_i$.
Calculating the non-zero values of $\mathbf{J}$ requires backpropagating from the residuals through the $\alpha$-blending (\cref{eq:alpha-blending}) and splat projection steps to the Gaussian parameters.
The tile-based rasterizer of 3DGS~\cite{kerbl3Dgaussians} performs this calculation using a \textit{per-pixel} parallelization.
That is, every thread handles one ray, stepping backwards along all splats that this ray hit.
We found that this parallelization is too slow for an efficient PCG implementation.
The reason is the repetition of the ray marching: per PCG iteration we do it once for $\mathbf{u}_i$ and once for $\mathbf{g}_i$.
As a consequence, the same intermediate $\alpha$-blending states (i.e., $T_s$, $\frac{\partial c}{\partial \alpha_s}$, $\frac{\partial c}{\partial c_s}$ for every splat $s$ along the ray) are re-calculated multiple (up to 18) times.

\mypar{Cache-driven parallelization}
\begin{figure*}
\includegraphics[width=\textwidth]{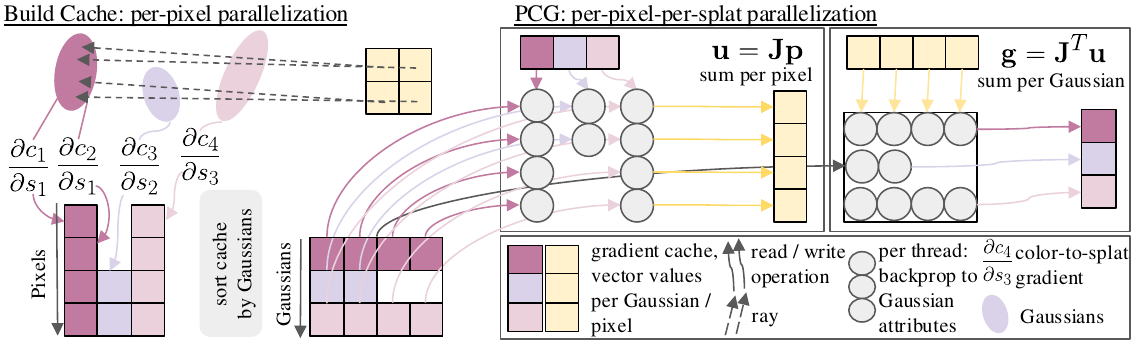}
\caption{
\textbf{Parallelization Strategy And Caching Scheme.}
We implement the PCG algorithm with efficient CUDA kernels, that use a gradient cache to calculate Jacobian-vector products.
Left: before PCG starts, we create the gradient cache following the \textit{per-pixel} parallelization of 3DGS~\cite{kerbl3Dgaussians}.
Afterwards, we sort the cache by Gaussians to ensure coalesced read accesses.
Right: the cache decouples splats along rays, which allows us to parallelize \textit{per-pixel-per-splat} when computing $\mathbf{u} = \mathbf{J} \mathbf{p}$ and $\mathbf{g} = \mathbf{J}^T \mathbf{u}$ during PCG.
}
\label{fig:parallelization}
\end{figure*}

We propose to change the parallelization to \textit{per-pixel-per-splat} (summarized in \cref{fig:parallelization}).
That is, one thread handles all residuals of one ray for one splat.
Each entry of $\mathbf{J}$ is the gradient from a residual $r$ (either of the L1 or SSIM terms) to a Gaussian parameter $x_i$.
Conceptually, this can be computed in three stages:
\begin{align}
\label{eq:backward-pass}
\frac{\partial r}{\partial x_i} = \frac{\partial r}{\partial c} \frac{\partial c}{\partial s} \frac{\partial s}{\partial x_i}
\end{align}
where $\frac{\partial r}{\partial c}$ denotes the gradient from the residual to the rendered color, $\frac{\partial c}{\partial s}$ from the color to the projected splat, and $\frac{\partial s}{\partial x_i}$ from the splat to the Gaussian parameter.
The first and last factors of \cref{eq:backward-pass} can be computed \textit{independently} for each residual and splat respectively, which allows for an efficient parallelization.
Similarly, we can calculate $\frac{\partial c}{\partial s}$ independently, if we have access to $T_s$ and $\frac{\partial c}{\partial \alpha_s}$.
Instead of looping over all splats along a ray multiple times, we cache these quantities once (\cref{fig:parallelization} left).
When calculating $\mathbf{u}_i$ or $\mathbf{g}_i$, we then read these values from the cache (\cref{fig:parallelization} right).
This allows us to parallelize over all splats in all pixels, which drastically accelerates the runtime.
The cache size is controlled by how many images (rays) we process in each PCG iteration and how many splats contribute to the final color along each ray.
We propose an efficient subsampling scheme that limits the cache size to the available budget.

3DGS uses the structural similarity index measure (SSIM) as loss term during optimization (\cref{eq:sgd-loss}).
In SSIM, the local neighborhood of every pixel gets convolved with Gaussian kernels to obtain the final per-pixel score~\cite{wang2004ssim}.
We calculate $\frac{\partial r}{\partial c}$ for the SSIM residuals by backpropagating the per-pixel scores to the center pixels (ignoring the contribution to other pixels in the local neighborhood).
This allows us to keep rays independent of each other thereby allowing for an efficient parallelization.
We implement it following the derivation of Zhao \etal~\cite{zhao2016loss}.

\mypar{Mapping of PCG to CUDA kernels}
We cache all gradients $\frac{\partial c}{\partial s}$ using the \texttt{\textcolor{blue}{buildCache}} operation.
Following the implementation of the differentiable rasterizer in 3DGS~\cite{kerbl3Dgaussians}, it uses the \textit{per-pixel} parallelization and calculates the gradient update $\mathbf{b} {=} -\mathbf{J}^T \mathbf{F}$.
For coalesced read and write accesses, we first store the cache sorted by pixels (\cref{fig:parallelization} left).
Afterwards, we re-sort it by Gaussians using the \texttt{\textcolor{blue}{sortCacheByGaussians}} kernel.
We use the Jacobi preconditioner $\mathbf{M}^{-1} {=} 1 / \text{diag}(\mathbf{J}^T \mathbf{J})$ and calculate it once using the \textit{per-pixel-per-splat} parallelization in the \texttt{\textcolor{blue}{diagJTJ}} kernel.
The inner PCG loop involves two kernels that are accelerated by our novel parallelization scheme.
First, \texttt{\textcolor{blue}{applyJ}} computes $\mathbf{u} {=} \mathbf{J} \mathbf{p}$, which we implement as a per-pixel sum aggregation.
Afterwards, \texttt{\textcolor{blue}{applyJT}} computes $\mathbf{g} {=} \mathbf{J}^T \mathbf{u}$.
This per-Gaussian sum can be efficiently aggregated using warp reductions.
We compute the remaining vector-vector terms of \cref{alg:pcg} directly in PyTorch~\cite{paszke2019pytorch}.
We refer to the supplementary material for more details.

\mypar{Image Subsampling Scheme}
Our cache consumes additional GPU memory.
For high resolution images in a dense reconstruction setup, the number of rays and thus the cache size can grow too large.
To this end, we split the images into batches and solve the normal equations independently, following ~\cref{eq:norm-eq}.
This allows us to store the cache only for one batch at a time.
Concretely, for $\text{n}_\text{b}$ batches, we obtain $\text{n}_\text{b}$ update vectors and combine them in a weighted mean:
\begin{align}
\label{eq:combine-pcg}
\Delta = \sum_{i=1}^{\text{n}_\text{b}} \frac{\mathbf{M}_i \Delta_i}{\sum_{k=1}^n \mathbf{M}_k}
\end{align}
where we use the inverse of the PCG preconditioner $\mathbf{M}_i {=} \text{diag}(\mathbf{J}_i^T \mathbf{J}_i)$ as the weights.
We refer to the supplementary material for a derivation of the weights.
These weights balance the importance of update vectors across batches based on how much each Gaussian parameter contributed to the rendered colors in the respective images.
This subsampling scheme allows us to control the cache size relative to the number of images in a batch.
In practice, we choose batch sizes of 25-70 images and up to $\text{n}_\text{b} {=} 4$ batches per LM iteration.
We either select the images at random or, if the scene was captured along a smooth trajectory, in a strided fashion to maximize scene coverage in all batches.

\subsection{3DGS Optimization In Two Stages}
\label{subsec:two-stage}
\begin{figure}
\includegraphics[width=0.5\textwidth]{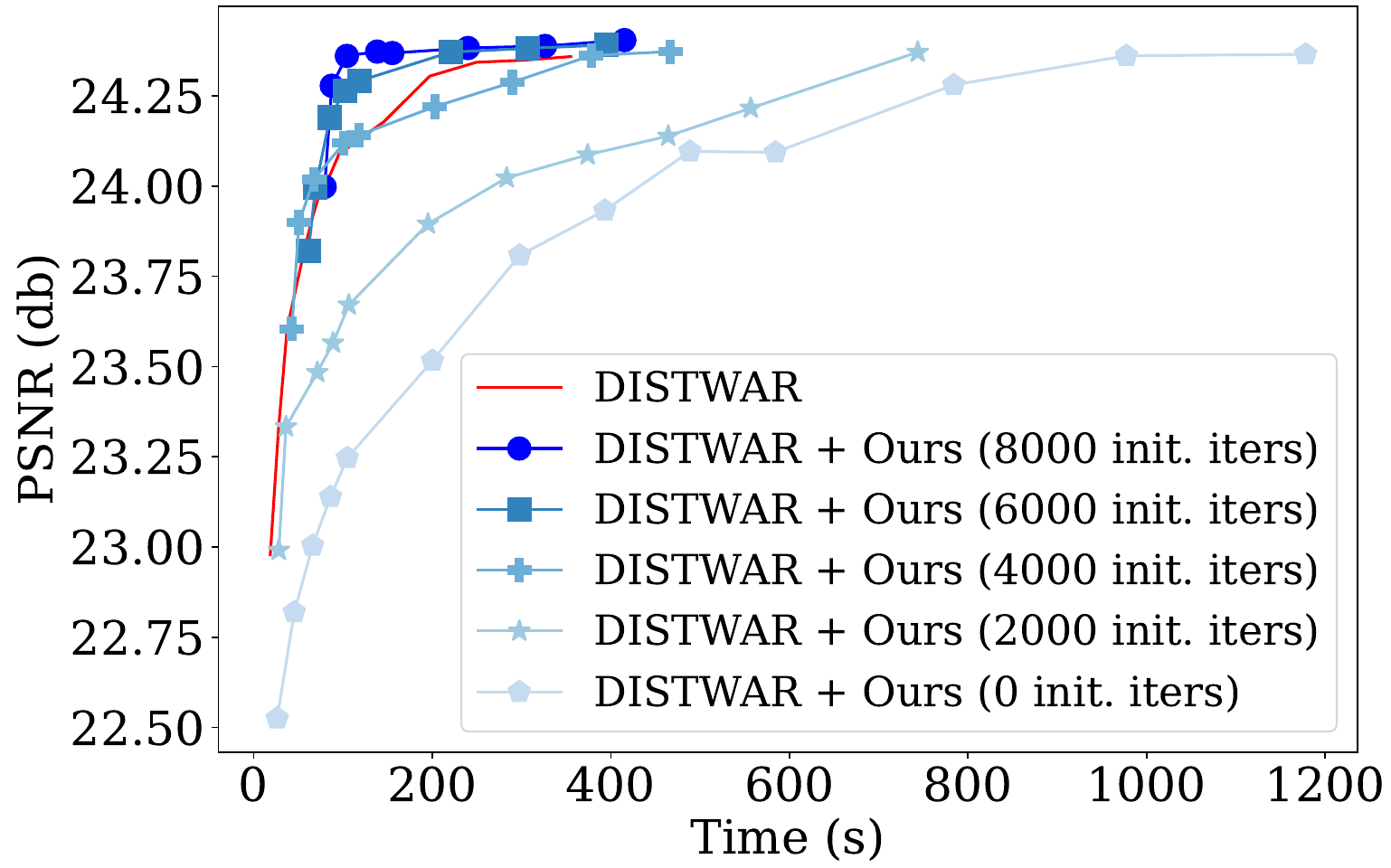}
\vspace{-7mm}
\caption{
\textbf{Comparison of initialization iterations.}
In our first stage, we initialize the Gaussians with gradient descent for $\text{K}$ iterations, before finetuning with our LM optimizer.
After $\text{K} {=} 6000$ or $\text{K} {=} 8000$ iterations, our method converges faster than the baseline.
With less iterations, pure LM is slower, which highlights the importance of our two stage approach.
Results reported on the \scene{GARDEN} scene from MipNeRF360~\cite{mildenhall2021nerf} without densification.
}
\label{fig:no-densification}
\end{figure}

Our pipeline utilizes the LM optimizer in the second stage of 3DGS optimization (see \cref{fig:pipeline}).
Before that, we run the ADAM optimizer to obtain an initialization of the Gaussian parameters.
We compare this against running our LM optimizer directly on the Gaussian initialization obtained from the SfM point cloud (following \cite{kerbl3Dgaussians}).
\cref{fig:no-densification} shows, that our LM converges faster for better initialized Gaussians and eventually beats pure ADAM. 
In contrast, running it directly on the SfM initialization is slower.
This demonstrates that quasi second-order solvers like ours are well-known to be more sensitive to initialization.
In other words, gradient descent makes rapid progress in the beginning, but needs more time to converge to final Gaussian parameters.
The additional compute overhead of our LM optimization is especially helpful to converge more quickly.
This motivates us to split the method in two stages.
It also allows us to complete the densification of the Gaussians before employing the LM optimizer, which simplifies the implementation.

\section{Results}
\label{sec:results}

\mypar{Baselines}
We compare our LM optimizer against ADAM in multiple reference implementations of 3DGS.
This shows, that our method is compatible with other runtime improvements.
In other words, we can swap out the optimizer and retain everything else.
Concretely, we compare against the original 3DGS~\cite{kerbl3Dgaussians}, its reimplementation ``gsplat''~\cite{ye2024gsplatopensourcelibrarygaussian}, and DISTWAR~\cite{durvasula2023distwar}.
Additionally, we compare against Taming-3DGS~\cite{mallick2024taming} by utilizing their ``budgeted'' approach as the fastest baseline in terms of runtime.
We run all baselines for 30K iterations with their default hyperparameters.

\mypar{Datasets / Metrics}
We benchmark our runtime improvements on three established datasets: Tanks\&Temples~\cite{knapitsch2017tanks}, Deep Blending~\cite{hedman2018deep}, and MipNeRF360~\cite{barron2022mip}.
These datasets contain in total 13 scenes that cover bounded indoor and unbounded outdoor environments.
We fit all scenes for every method on the same NVIDIA A100 GPU using the train/test split as proposed in the original 3DGS~\cite{kerbl3Dgaussians} publication.
To measure the quality of the reconstruction, we report peak signal-to-noise ratio (PSNR), structural similarity (SSIM), and perceptual similarity (LPIPS)~\cite{zhang2018perceptual} averaged over all test images.
Additionally, we report the optimization runtime and the maximum amount of consumed GPU memory.

\mypar{Implementation Details}
For our main results, we run the first stage for 20K iterations with the default hyperparameters of the respective baseline.
The densification is completed after 15K iterations.
Afterwards, we only have to run 5 LM iterations with 8 PCG iterations each to converge on all scenes.
This showcases the efficiency of our optimizer.
Since the image resolutions are different for every dataset, we select the batch-size and number of batches such that the consumed memory for caching is similar.
We select 25 images in 4 batches for MipNeRF360~\cite{barron2022mip}, 25 images in 3 batches for Deep Blending~\cite{hedman2018deep}, and 70 images in 3 batches for Tanks\&Temples~\cite{knapitsch2017tanks}.
We constrain the value range of $\lambda_\text{reg}$ for stable updates. %
We define it in $[1\mathrm{e}{-4}, 1\mathrm{e}{4}]$ for Deep Blending~\cite{hedman2018deep} and Tanks\&Temples~\cite{knapitsch2017tanks} and in the interval $[1\mathrm{e}{-4}, 1\mathrm{e}{-2}]$ for MipNeRF360~\cite{barron2022mip}.

\subsection{Comparison To Baselines}
\begin{table*}
  \centering
  \setlength\tabcolsep{2pt}
  \begin{tabular}{l | rrrr | rrrr | rrrr}
    \toprule
        \multirow{2}{*}{Method} & \multicolumn{4}{c}{MipNeRF-360~\cite{barron2022mip}} & \multicolumn{4}{c}{Tanks\&Temples~\cite{knapitsch2017tanks}} & \multicolumn{4}{c}{Deep Blending~\cite{hedman2018deep}} \\
        \cmidrule(l{2pt}r{2pt}){2-5} \cmidrule(l{2pt}r{2pt}){6-9} \cmidrule(l{2pt}r{2pt}){10-13}
    & SSIM$\uparrow$ & PSNR$\uparrow$ & LPIPS$\downarrow$ & Time (s)
    & SSIM$\uparrow$ & PSNR$\uparrow$ & LPIPS$\downarrow$ & Time (s)
    & SSIM$\uparrow$ & PSNR$\uparrow$ & LPIPS$\downarrow$ & Time (s) \\
    \midrule
    3DGS~\cite{kerbl3Dgaussians} & 0.813 & 27.40 & 0.218 & 1271 & 0.844 & 23.68 & 0.178 & 736 & 0.900 & 29.51 & 0.247 & 1222 \\
    + Ours & 0.813 & 27.39 & 0.221 & \textbf{972} & 0.845 & 23.73 & 0.182 & \textbf{663} & 0.903 & 29.72 & 0.247 & \textbf{951} \\
    \midrule
    DISTWAR~\cite{durvasula2023distwar} & 0.813 & 27.42 & 0.217 & 966 & 0.844 & 23.67 & 0.178 & 601 & 0.899 & 29.47 & 0.247 & 841 \\
    + Ours & 0.814 & 27.42 & 0.221 & \textbf{764} & 0.844 & 23.67 & 0.183 & \textbf{537} & 0.902 & 29.60 & 0.248 & \textbf{672} \\
    \midrule
    gsplat~\cite{ye2024gsplatopensourcelibrarygaussian} & 0.814 & 27.42 & 0.217 & 1064 & 0.846 & 23.50 & 0.179 & 646 & 0.904 & 29.52 & 0.247 & 919 \\
    + Ours & 0.814 & 27.42 & 0.221 & \textbf{818} & 0.844 & 23.68 & 0.183 & \textbf{414} & 0.902 & 29.58 & 0.249 & \textbf{716} \\
    \midrule
    Taming-3DGS~\cite{mallick2024taming} & 0.793 & 27.14 & 0.260 & 566 & 0.833 & 23.76 & 0.209 & 366 & 0.900 & 29.84 & 0.274 & 447 \\
    + Ours & 0.791 & 27.13 & 0.260 & \textbf{453} & 0.832 & 23.72 & 0.209 & \textbf{310} & 0.901 & 29.91 & 0.275 & \textbf{347} \\
    \midrule
    \bottomrule
  \end{tabular}
  \caption{
    \textbf{Quantitative comparison of our method and baselines.}
    By adding our method to baselines, we accelerate the optimization time by 20\% on average while achieving the same quality.
    We can combine our method with others, that improve runtime along different axes.
    This demonstrates that our method offers an orthogonal improvement, i.e., the LM optimizer can be plugged into many existing methods.
  }
  \label{tab:quant-comparison}
\end{table*}

\begin{figure*}
\centering
\setlength\tabcolsep{1pt}
\begin{tabular}{ccc}
\includegraphics[width=0.33\textwidth]{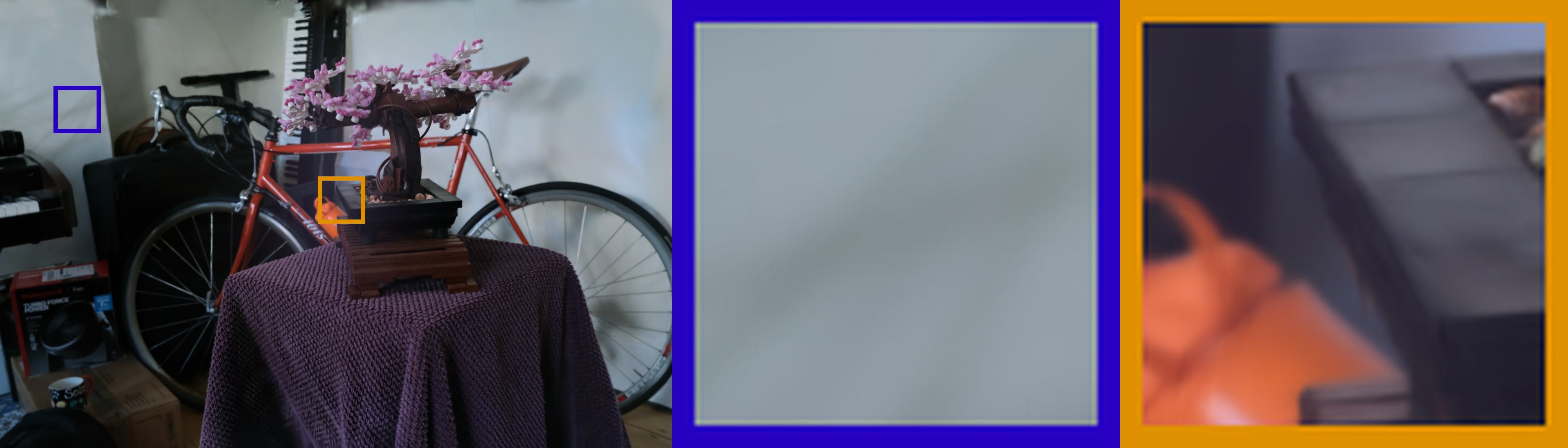} &
\includegraphics[width=0.33\textwidth]{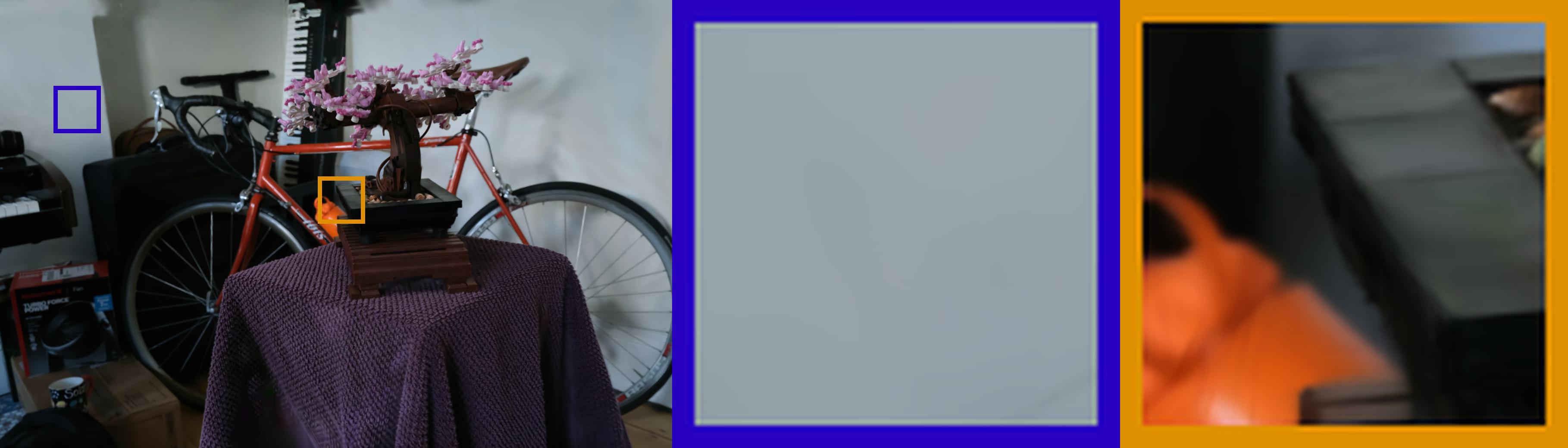} &
\includegraphics[width=0.33\textwidth]{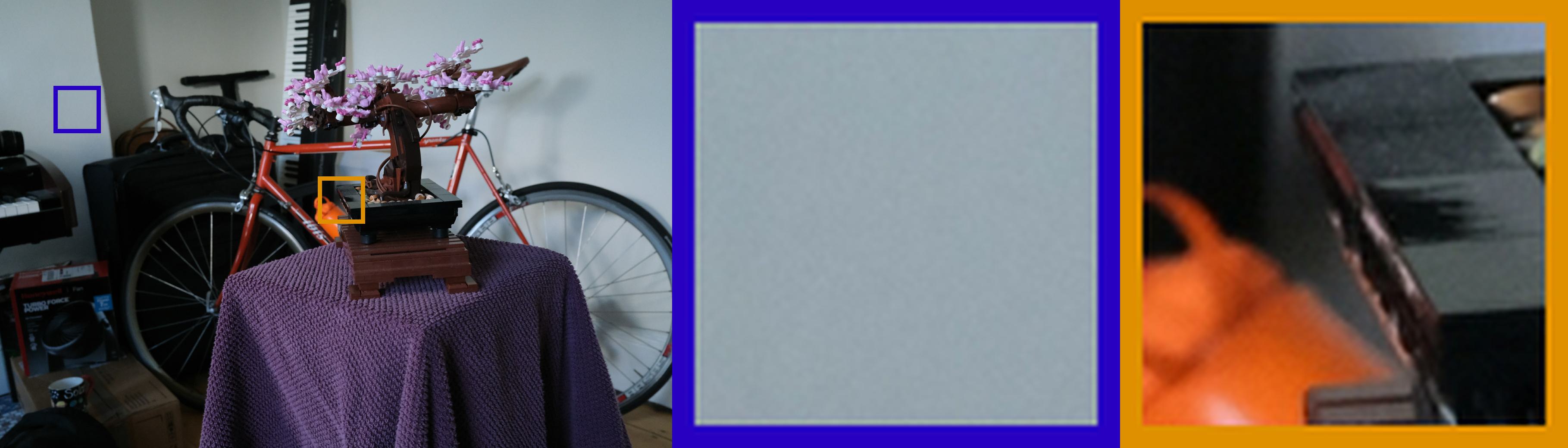} \\
3DGS~\cite{kerbl3Dgaussians} after 814s & 3DGS + Ours after 794s & Ground-Truth Images \\

\includegraphics[width=0.33\textwidth]{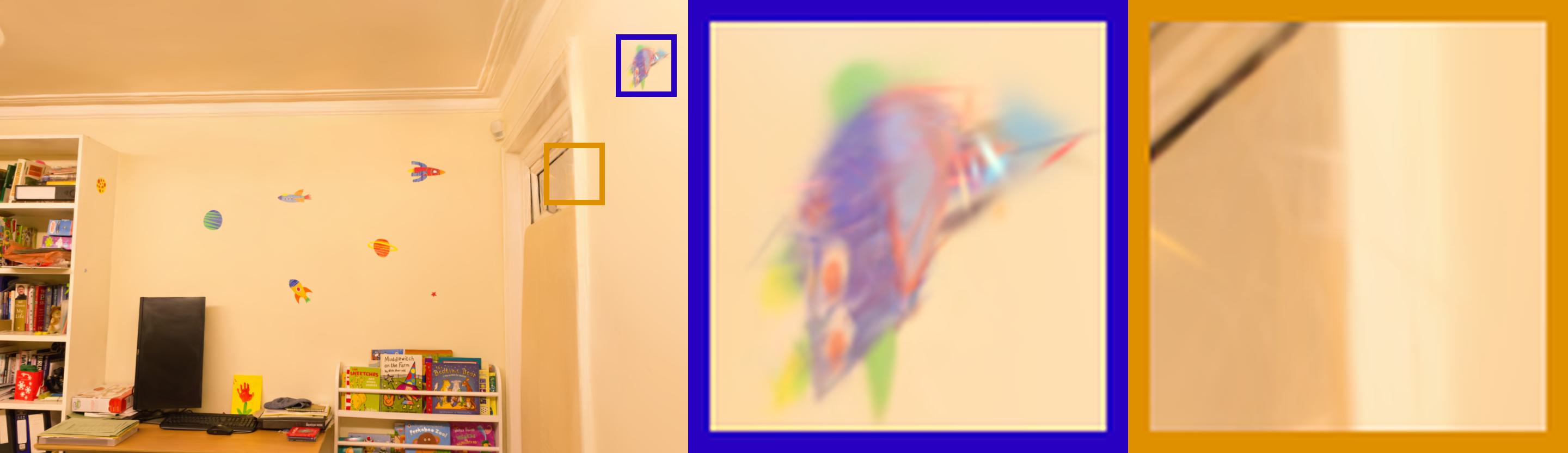} &
\includegraphics[width=0.33\textwidth]{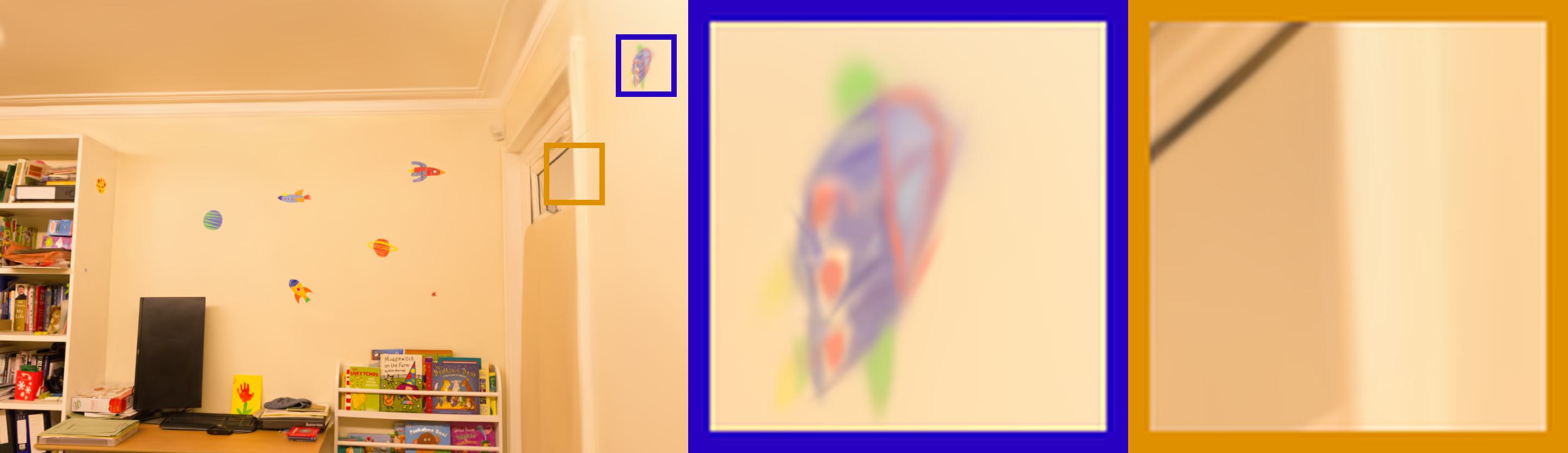} &
\includegraphics[width=0.33\textwidth]{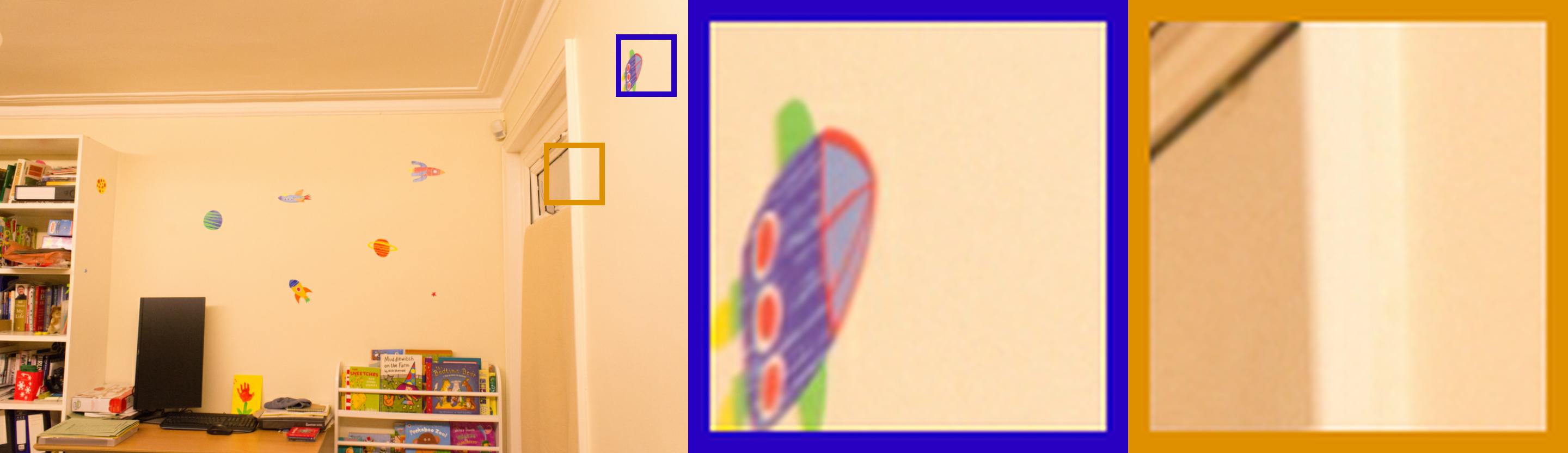} \\
Taming-3DGS~\cite{mallick2024taming} after 328s & Taming-3DGS + Ours after 324s & Ground-Truth Images \\

\includegraphics[width=0.33\textwidth]{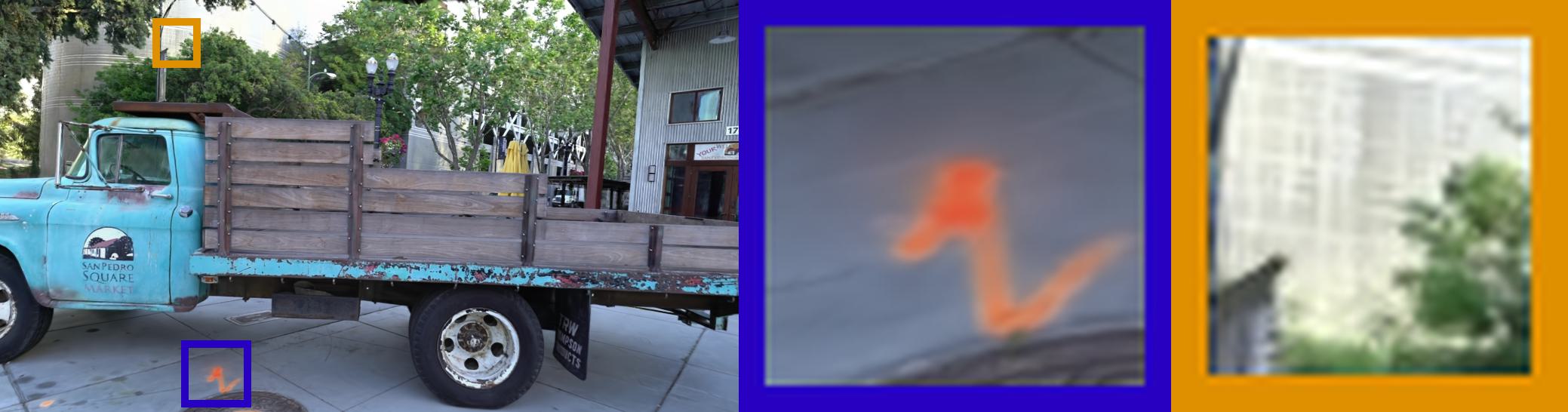} &
\includegraphics[width=0.33\textwidth]{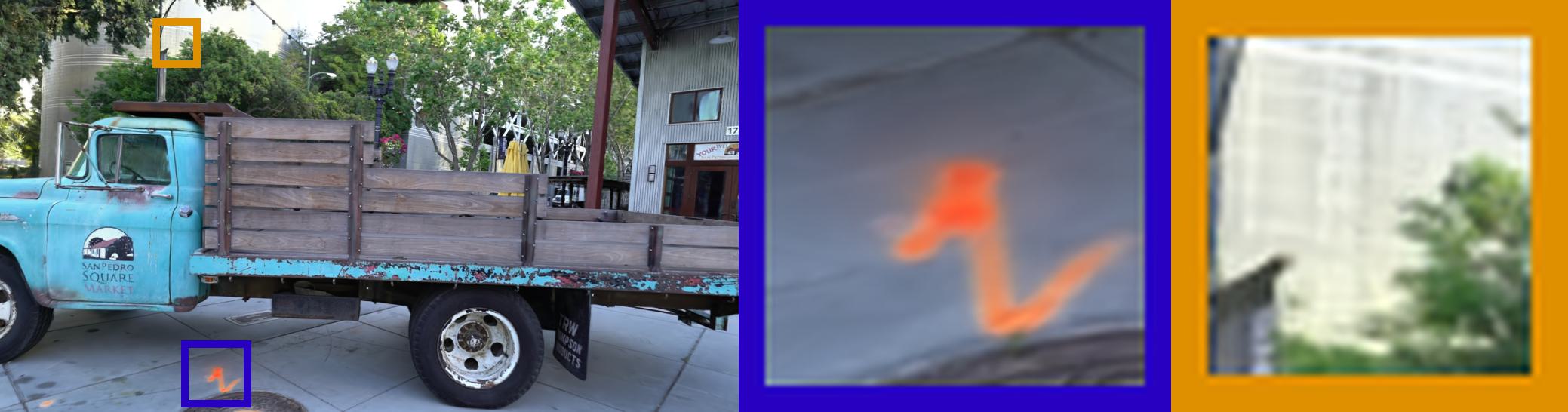} &
\includegraphics[width=0.33\textwidth]{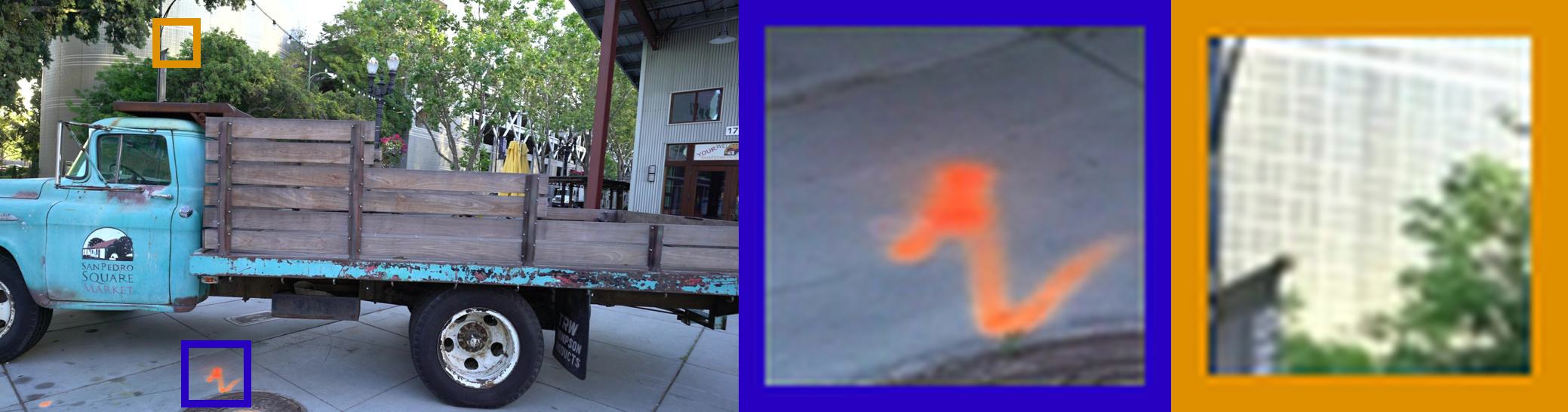} \\
gsplat~\cite{ye2024gsplatopensourcelibrarygaussian} after 453s & gsplat + Ours after 447s & Ground-Truth Images \\

\includegraphics[width=0.33\textwidth]{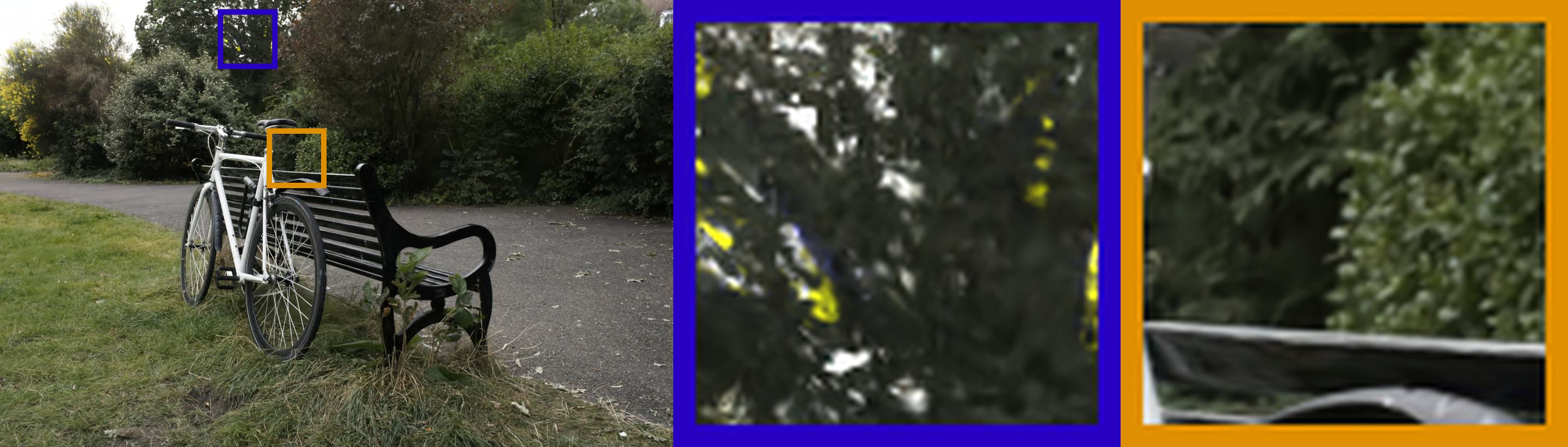} &
\includegraphics[width=0.33\textwidth]{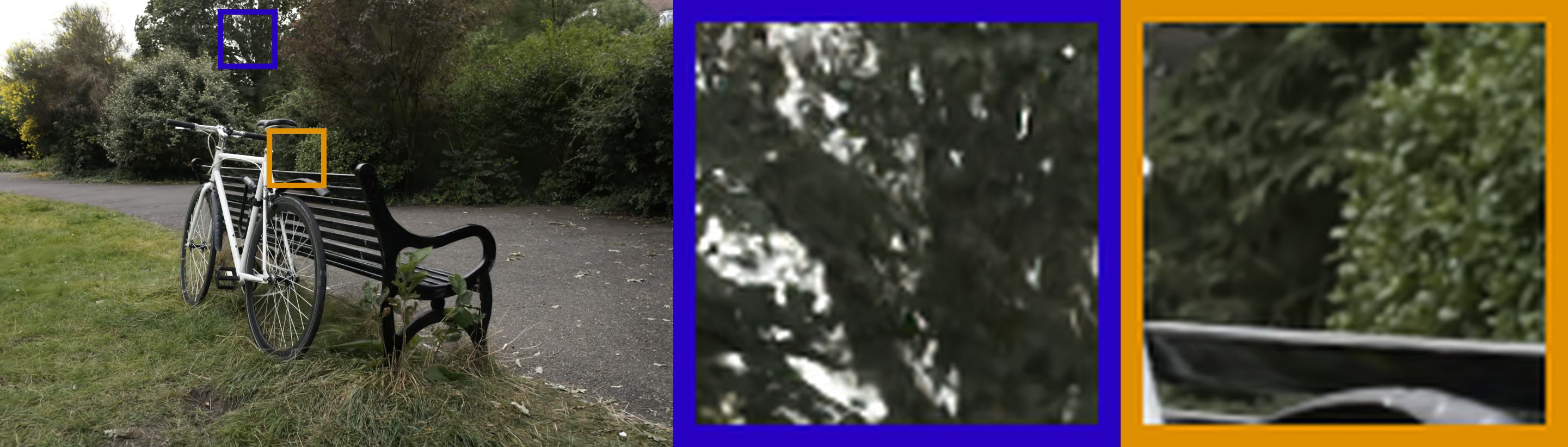} &
\includegraphics[width=0.33\textwidth]{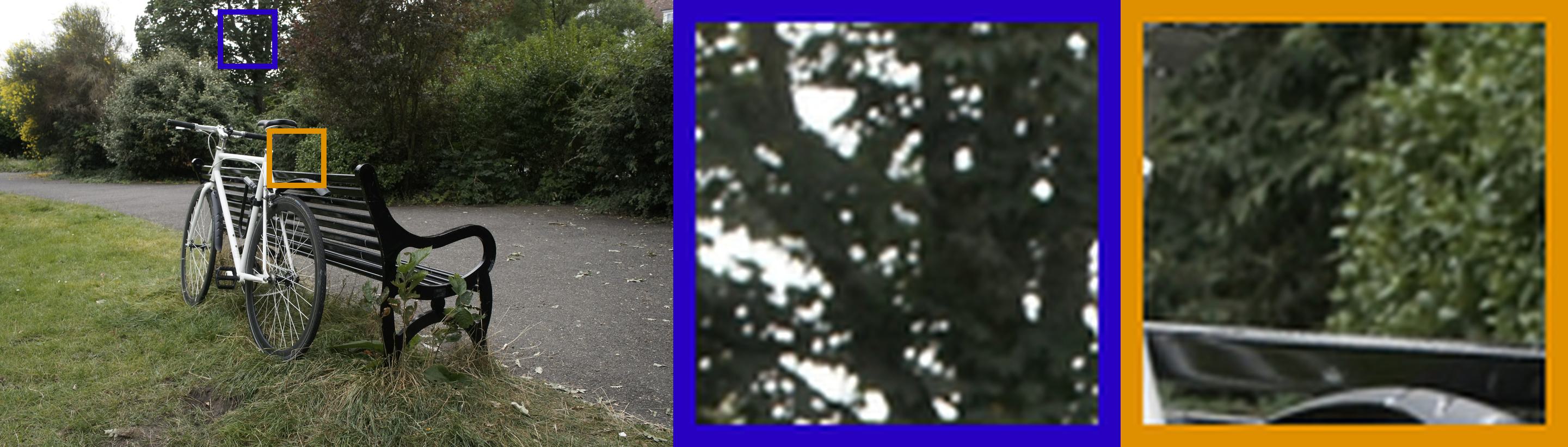} \\
DISTWAR~\cite{durvasula2023distwar} after 978s & DISTWAR + Ours after 971s & Ground-Truth Images \\

\end{tabular}
\caption{
\textbf{Qualitative comparison of our method and baselines.}
We compare rendered test images after similar optimization time.
All baselines converge faster when using our LM optimizer, which shows in images with fewer artifacts and more accurate brightness / contrast.
}
\label{fig:qual-comp}
\end{figure*}

We report our main quantitative results in~\cref{tab:quant-comparison}.
Our LM optimizer can be added to all baseline implementations and accelerates the optimization runtime by 20\% on average.
The reconstructions show similar quality across all metrics and datasets, highlighting that our method arrives at similar local minima, just faster.
We also provide a per-scene breakdown of these results in the supplementary material.
On average our method consumes 53 GB of GPU memory on all datasets.
In contrast, the baselines do not use an extra cache and only require between 6-11 GB of memory.
This showcases the runtime-memory tradeoff of our approach.

We visualize sample images from the test set in~\cref{fig:qual-comp} for both indoor and outdoor scenarios.
After the same amount of optimization runtime, our method is already converged whereas the baselines still need to run longer.
As a result, the baselines still contain suboptimal Gaussians, which results in visible artifacts in rendered images.
In comparison, our rendered images more closely resemble the ground truth with more accurate brightness / contrast and texture details.

\subsection{Ablations}

\mypar{Is the L1/SSIM objective important?}
We utilize the same objective in our LM optimizer as in the original 3DGS implementation, namely the L1 and SSIM loss terms (\cref{eq:sgd-loss}).
Since LM energy terms are defined as a sum of squares, we adopt the square root formulation of these loss terms to arrive at an identical objective (\cref{eq:gn-loss}).
We compare this choice against fitting the Gaussians with only an L2 loss, that does not require taking a square root.
Concretely, we compare the achieved quality and runtime of LM against ADAM for both the L2 loss and the L1 and SSIM losses.
As can be seen in \cref{tab:abl-l2-l1-ssim}, we achieve faster convergence and similar quality in both cases.
However, the achieved quality is inferior for both LM and ADAM when only using the L2 loss.
This highlights the importance of the L1 and SSIM loss terms and why we adopt them in our method as well.
We show in the supplementary material, that computing these loss terms instead of the simpler L2 residuals does not negatively impact the efficiency of our CUDA kernels.

\mypar{How many images per batch are necessary?}
The key hyperparameters in our model are the number of images in a batch and how many batches to choose for every LM iteration (\cref{subsec:gsgn-kernels}).
This controls the runtime of one iteration and how much GPU memory our optimizer consumes.
We compare different numbers of images in \cref{tab:abl-img-init} on the NeRF-Synthetic~\cite{mildenhall2021nerf} dataset in a single batch per LM iteration, i.e., $\text{n}_\text{b} {=} 1$.
Using the full dataset (100 images) produces the best results.
Decreasing the number of images in a batch results in only slightly worse quality, but also yields faster convergence and reduces GPU memory consumption linearly down to 15GB for 40 images.
This demonstrates that subsampling images does not negatively impact the convergence of the LM optimizer in our task.

\mypar{Are we better than multi-view ADAM?}
Our method converges with fewer iterations than baselines.
Concretely, we require only 5-10 additional LM iterations after the initialization, whereas ADAM runs for another 10K iterations.
We increase the batch-size (number of images) for the baselines, such that the same number of multi-view constraints are observed for the respective update steps. 
However, as can be seen in \cref{tab:abl-mv-constraints}, the achieved quality is worse for ADAM after the same number of iterations.
When running for more iterations, ADAM eventually converges to similar quality, but needs more time.
This highlights the efficiency of our optimizer: since we solve the normal equations in \cref{eq:gn-loss}, one LM iteration makes a higher quality update step than ADAM which only uses the gradient direction.

\subsection{Runtime Analysis}

\begin{figure}
\includegraphics[width=0.5\textwidth]{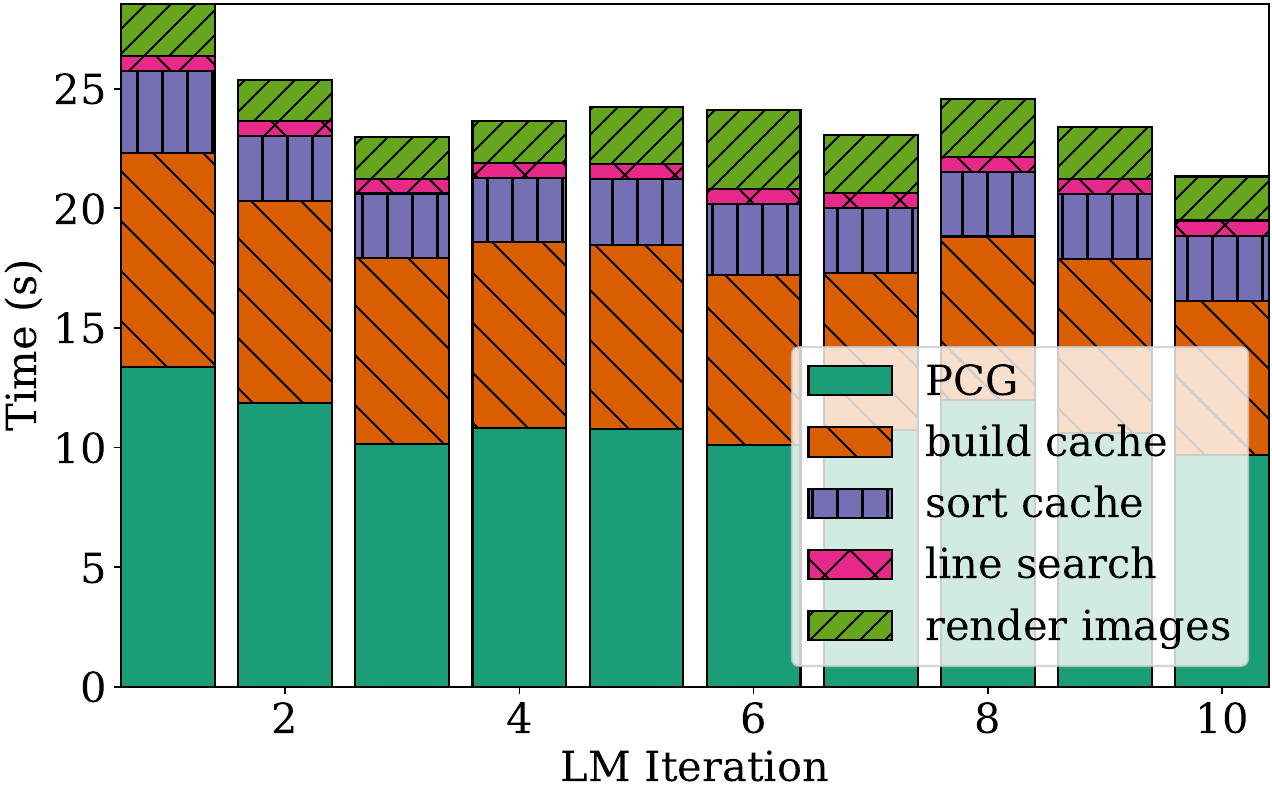}
\caption{
\textbf{Runtime Analysis.}
One iteration of our LM optimizer is dominated by solving PCG and building the cache.
Measured on the \scene{GARDEN} scene from Mip-NeRF360~\cite{barron2022mip} after densification.
}
\vspace{-3mm}
\label{fig:runtime}
\end{figure}

We analyze the runtime of our LM optimizer across multiple iterations in \cref{fig:runtime}.
The runtime is dominated by solving \cref{eq:norm-eq} with PCG and building the cache (\cref{subsec:gsgn-kernels}).
Sorting the cache, rendering the selected images, and the line search (\cref{eq:line-search}) are comparatively faster.
During PCG, we run the \texttt{\textcolor{blue}{applyJ}} and \texttt{\textcolor{blue}{applyJT}} kernels up to 9 times, parallelizing \textit{per-pixel-per-splat}.
In contrast, we run the \texttt{\textcolor{blue}{buildCache}} kernel once, parallelizing \textit{per-pixel}, which is only marginally faster.
This shows the advantage of our proposed parallelization scheme: the same Jacobian-vector product runs much faster.
We also provide a detailed profiling analysis of our kernels in the supplementary material.

\subsection{Limitations}
\begin{table}
  \centering
  \setlength\tabcolsep{2pt}
  \begin{tabular}{l | rrrr}
    \toprule
    Method & SSIM$\uparrow$ & PSNR$\uparrow$ & LPIPS$\downarrow$ & Time (s) \\
    \midrule
    3DGS~\cite{kerbl3Dgaussians} (L1/SSIM) & 0.862 & 27.23 & \textbf{0.108} & 1573 \\
    3DGS + Ours (L1/SSIM) & \textbf{0.863} & \textbf{27.29} & 0.110 & \textbf{1175} \\
    \midrule
    3DGS~\cite{kerbl3Dgaussians} (L2) & 0.854 & 27.31 & 0.117 & 1528 \\
    3DGS + Ours (L2) & \textbf{0.857} & \textbf{27.48} & \textbf{0.114} & \textbf{1131} \\
    \bottomrule
  \end{tabular}
  \caption{
    \textbf{Ablation of objective.}
    We compare using the L1/SSIM losses against the L2 loss.
    For both, 3DGS~\cite{kerbl3Dgaussians} optimized with ADAM and combined with ours, we achieve better results with the L1/SSIM objective.
    In both cases, our method accelerates the convergence.
    Results on the \scene{GARDEN} scene from MipNeRF360~\cite{barron2022mip}.
  }
  \label{tab:abl-l2-l1-ssim}
\end{table}

\begin{table}
  \centering
  \setlength\tabcolsep{2pt}
  \begin{tabular}{l | rrrrr}
    \toprule
    Batch Size & SSIM$\uparrow$ & PSNR$\uparrow$ & LPIPS$\downarrow$ & Time (s) & Mem (Gb) \\
    \midrule
    100 & \textbf{0.969} & \textbf{33.77} & \textbf{0.030} & 242 & 32.5 \\
    80 & \textbf{0.969} & 33.73 & 0.031 & 233 & 29.8 \\
    60 & 0.968 & 33.69 & 0.031 & 223 & 22.6 \\
    40 & 0.967 & 33.51 & 0.032 & 212 & 15.4 \\
    \bottomrule
  \end{tabular}
  \caption{
    \textbf{Ablation of batch-size.}
    Selecting fewer images per LM iteration reduces runtime and consumed GPU memory, while only slightly impacting quality.
    This demonstrates that image subsampling (\cref{subsec:gsgn-kernels}) is compatible with LM in our task.
    Results obtained after initialization with 3DGS~\cite{kerbl3Dgaussians} and with $\text{n}_\text{b} {=} 1$.
  }
  \label{tab:abl-img-init}
\end{table}

\begin{table}
  \centering
  \setlength\tabcolsep{2pt}
  \begin{tabular}{l | cc | rr}
    \toprule
        Method & Iterations & Batch-Size & Time (s) & PSNR$\uparrow$ \\
    \midrule
    3DGS~\cite{kerbl3Dgaussians} & 10,000 & 1 & 1222 & 29.51 \\
    3DGS~\cite{kerbl3Dgaussians} & 50 & 75 & 962 & 29.54 \\
    3DGS~\cite{kerbl3Dgaussians} & 130 & 75 & 1193 & 29.68 \\
    + Ours & 5 & 75 & \textbf{951} & 29.72 \\
    \midrule
    DISTWAR~\cite{durvasula2023distwar} & 10,000 & 1 & 841 & 29.47 \\
    DISTWAR~\cite{durvasula2023distwar} & 50 & 75 & 681 & 29.49 \\
    DISTWAR~\cite{durvasula2023distwar} & 130 & 75 & 814 & 29.58 \\
    + Ours & 5 & 75 & \textbf{672} & 29.60 \\
    \midrule
    gsplat~\cite{ye2024gsplatopensourcelibrarygaussian} & 10,000 & 1 & 919 & 29.52 \\
    gsplat~\cite{ye2024gsplatopensourcelibrarygaussian} & 50 & 75 & 724 & 29.53 \\
    gsplat~\cite{ye2024gsplatopensourcelibrarygaussian} & 130 & 75 & 892 & 29.56 \\
    + Ours & 5 & 75 & \textbf{716} & 29.58 \\
    \midrule
    Taming-3DGS~\cite{mallick2024taming} & 10,000 & 1 & 447 & 29.84 \\
    Taming-3DGS~\cite{mallick2024taming} & 50 & 75 & 328 & 29.86 \\
    Taming-3DGS~\cite{mallick2024taming} & 130 & 75 & 391 & 29.91 \\
    + Ours & 5 & 75 & \textbf{347} & 29.91 \\
    \bottomrule
  \end{tabular}
    \caption{
    \textbf{Analysis of multi-view constraints.}
    We obtain higher quality update steps from our LM optimization and need fewer iterations to converge.
    Using equally many images in a batch, baselines using ADAM still require more iterations and runtime to reach similar quality.
    Results averaged on DeepBlending~\cite{hedman2018deep}.
    }
  \label{tab:abl-mv-constraints}
\end{table}

By replacing ADAM with our LM scheme, we accelerate the 3DGS convergence speed by 20\% on average for all datasets and baselines.
However, some drawbacks remain.
First, our approach requires more GPU memory than baselines, due to our gradient cache (\cref{subsec:gsgn-kernels}).
Depending on the number and resolution of images, this might require additional CPU offloading of cache parts to run our method on smaller GPUs.
Following Mallick \etal~\cite{mallick2024taming}, one can further reduce the cache size by storing the gradients $\frac{\partial c}{\partial s}$ only for every 32nd splat along a ray and re-doing the $\alpha$-blending in these local windows.
Second, our two-stage approach relies on ADAM for the densification.
3DGS~\cite{kerbl3Dgaussians} densifies Gaussians up to 140 times, which is not easily transferable to the granularity of only 5-10 LM iterations.
Instead, one could explore and integrate recent alternatives~\cite{kheradmand20243d, bulo2024revising, lu2024turbo}.

\section{Conclusion}
\label{sec:conclusion}

We have presented \OURS, a method that accelerates the reconstruction of 3D Gaussian-Splatting \cite{kerbl3Dgaussians} by replacing the ADAM optimizer with a tailored Levenberg-Marquardt (LM) (\cref{subsec:gsgn-theory}).
We show that with our data parallelization scheme we can efficiently solve the normal equations with PCG in custom CUDA kernels (\cref{subsec:gsgn-kernels}).
Employed in a two-stage approach (\cref{subsec:two-stage}), this leads to a 20\% runtime acceleration compared to baselines.
We further demonstrate that our approach is agnostic to other methods \cite{durvasula2023distwar, ye2024gsplatopensourcelibrarygaussian, mallick2024taming}, which further improves the optimization runtime; i.e., we can easily combine our proposed optimizer with faster 3DGS methods. %
Overall, we believe that the ability of faster 3DGS reconstructions with our method will open up further research avenues like \cite{lan20253dgs} and make 3DGS more practical across a wide range of real-world applications.\looseness-1

\section{Acknowledgements}
\label{sec:ack}

This project was funded by a Meta sponsored research agreement. In addition, the project was supported by the ERC Starting Grant Scan2CAD (804724) as well as the German Research Foundation (DFG) Research Unit ``Learning and Simulation in Visual Computing''.
We thank Justin Johnson for the helpful discussions in an earlier project with a similar direction and Peter Kocsis for the helpul discussions about image subsampling.
We also thank Angela Dai for the video voice-over.

{
    \small
    \bibliographystyle{ieeenat_fullname}
    \bibliography{main}
}

\clearpage
\maketitlesupplementary
\appendix

\section{More Details About CUDA Kernel Design}
\label{subsec:add-detail-kernels}
We introduce the necessary CUDA kernels to calculate the PCG algorithm in \cref{subsec:gsgn-kernels}.
In this section, we provide additional implementation details.

\subsection{Parallelization Pattern}
We implement the \textit{per-pixel-per-splat} parallelization pattern in our CUDA kernels by reading subsequent entries from the gradient cache in subsequent threads.
This makes reading cache values perfectly coalesced and therefore minimizes the overhead caused by the operation.
The gradient cache is sorted over Gaussians, which means that subsequent entries refer to different rays (pixels) that saw this projected Gaussian (splat).
In general, one thread handles all residuals corresponding to the respective pixel, i.e., many computations are shared across color channels and are therefore combined.

\subsection{Design Of \texttt{buildCache} And \texttt{applyJT}}
The necessary computations for the \texttt{buildCache} and \texttt{applyJT} steps in PCG (see \cref{alg:pcg}) are split across three kernels.
This follows the original design of the 3DGS differentiable rasterizer~\cite{kerbl3Dgaussians}.
In both cases, we calculate the Jacobian-vector product of the form $\mathbf{g} {=} \mathbf{J}^T \mathbf{u}$ where $\mathbf{J} {\in} \mathbb{R}^{NxM}$ is the Jacobian matrix of $N$ residuals and $M$ Gaussian parameters and $\mathbf{u} {\in} \mathbb{R}^N$ is an input vector.
The $k$-th element in the output vector is calculated as:
\begin{align}
\label{eq:apply-jt}
\mathbf{g}_k = \sum_{i=0}^{N} \frac{\partial \mathbf{r}_i}{\partial \mathbf{x}_k} \mathbf{u}_i = \frac{\partial \mathbf{y}_k}{\partial \mathbf{x}_k} \sum_{i=0}^{N} \frac{\partial \mathbf{r}_i}{\partial \mathbf{y}_k} \mathbf{u}_i
\end{align}
where $\mathbf{r}_i$ is the $i$-th residual and $\mathbf{x}_k$ is the $k$-th Gaussian parameter.
In other words, $\mathbf{g}_k$ is a sum over all residuals for the $k$-th Gaussian parameter.
Following the chain-rule, it is possible to split up the gradient $\frac{\partial \mathbf{r}_i}{\partial \mathbf{x}_k} {=} \frac{\partial \mathbf{r}_i}{\partial \mathbf{y}_k} \frac{\partial \mathbf{y}_k}{\partial \mathbf{x}_k}$.
We can use this to split the computation across three smaller kernels, where only the first needs to calculate the sum over all residuals: $\sum_{i=0}^{N} \frac{\partial \mathbf{r}_i}{\partial \mathbf{y}_k} \mathbf{u}_i$.
This sum is the main bottleneck for the kernel implementation, since it needs to be implemented atomically (i.e., multiple threads write to the same output position).
The other kernels then only calculate the remaining steps by parallelizing over Gaussians.
We slightly abuse notation and denote with $\mathbf{y}_k$ the 2D mean, color, and opacity attributes of the $k$-th projected Gaussian.
That is, the first kernel sums up the partial derivatives to each of these attributes in separate vectors.
In the following, we add the suffixes \texttt{\_p1, \_p2, \_p3} to denote the three kernels for the respective operation (where \texttt{\_p1} refers to the kernel that calculates the sum over residuals).

The \texttt{buildCache\_p1} utilizes the original \textit{per-pixel} parallelization, whereas the \texttt{applyJT\_p1} kernel use the gradient cache and our proposed \textit{per-pixel-per-splat} parallelization pattern.
The gradient cache is sorted over Gaussians, i.e., subsequent entries correspond to different rays of the same splat.
This allows us to efficiently implement the sum by first performing a segmented warp reduce and then only issuing one \texttt{atomicAdd} statement per warp.

\subsection{Design Of \texttt{applyJ} And \texttt{diagJTJ}}
In contrast, the \texttt{applyJ} and \texttt{diagJTJ} computations cannot be split up into smaller kernels.
Concretely, the \texttt{applyJ} kernel calculates $\mathbf{u} {=} \mathbf{J} \mathbf{p}$ with $\mathbf{p} {\in} \mathbb{R}^M$. The $k$-th element in the output vector is calculated as:
\begin{align}
\label{eq:apply-j}
\mathbf{u}_k = \sum_{i=0}^{M} \frac{\partial \mathbf{r}_k}{\partial \mathbf{x}_i} \mathbf{p}_i = \sum_{i=0}^{N} \frac{\partial \mathbf{r}_k}{\partial \mathbf{y}_i} \frac{\partial \mathbf{y}_i}{\partial \mathbf{x}_i} \mathbf{p}_i
\end{align}
In other words, $\mathbf{u}_k$ is a sum over all Gaussian attributes for the $k$-th residual.
Similarly, the \texttt{diagJTJ} kernel calculates $\mathbf{M} = \text{diag}(\mathbf{J}^T \mathbf{J}) \in \mathbb{R}^M$.
The $k$-th element in the output vector is calculated as:
\begin{align}
\label{eq:diag-jtj}
\mathbf{M}_k = \sum_{i=0}^{N} (\frac{\partial \mathbf{r}_i}{\partial \mathbf{x}_k})^2 = \sum_{i=0}^{N} (\frac{\partial \mathbf{r}_i}{\partial \mathbf{y}_k} \frac{\partial \mathbf{y}_k}{\partial \mathbf{x}_k})^2
\end{align}
In both cases it is not possible to move part of the gradients outside of the sum.
As a consequence, both \texttt{applyJ} and \texttt{diagJTJ} are implemented as one kernel, where each thread directly calculates the final partial derivatives to all Gaussian attributes.
This slightly increases the number of required registers and the runtime compared to the \texttt{applyJT} kernel (see \cref{tab:profiler-analysis}).
The \texttt{diagJTJ} kernel makes use of the same segmented warp reduce as \texttt{applyJT\_p1} for efficiently summing up the squared partial derivatives.
The \texttt{applyJ} kernel first sums up over all Gaussian attributes within each thread separately.
Then, we only issue one \texttt{atomicAdd} statement for each residual per thread.

The \texttt{applyJ} kernel requires the input vector $\mathbf{p}$ to be sorted per Gaussian to make reading from it coalesced.
That is: $\mathbf{p} {=} [x_1^a, ..., x_1^z, ..., x_M^a, ..., x_M^z]^T$, where $x_k^a$ is the value corresponding to the $a$-th parameter of the $k$-th Gaussian.
In total, each Gaussian consists of 59 parameters: 11 for position, rotation, scaling, and opacity and 48 for all Spherical Harmonics coefficients of degree 3.
In contrast, all other kernels require the output vector to be sorted per attribute to make writing to it coalesced.
That is: $\mathbf{q} {=} [x_1^a, ..., x_M^a, ..., x_1^z, ..., x_M^z]^T$.
We use the structure of $\mathbf{q}$ for all other vector-vector calculations in \cref{alg:pcg} as well.
Whenever we call the \texttt{applyJ} kernel, we thus first call the \texttt{sortX} kernel that restructures $\mathbf{q}$ to the layout of $\mathbf{p}$.

\subsection{Precomputation Of Residual-To-Pixel Weights}
We adopt the square root formulation of the residuals in our energy formulation (see \cref{eq:gn-loss}).
We efficiently precompute the contribution of the square root to the partial derivatives of \cref{eq:apply-jt}, \cref{eq:apply-j}, and \cref{eq:diag-jtj}.
In the following, we divide the partial derivatives from the $i$-th residual to the $k$-th Gaussian attribute into two stages:
\begin{align}
\label{eq:res-sq-root-weights}
\frac{\partial \mathbf{r}_i}{\partial \mathbf{x}_k} = \frac{\partial \mathbf{r}_i}{\partial \mathbf{c}_i} \frac{\partial \mathbf{c}_i}{\partial \mathbf{x}_k}
\end{align}
where $\frac{\partial \mathbf{r}_i}{\partial \mathbf{c}_i}$ goes from the residual to the rendered pixel color and $\frac{\partial \mathbf{c}_i}{\partial \mathbf{x}_k}$ from the pixel color to the Gaussian attribute.
Since we adopt the L1 and SSIM loss terms and take their square root, the terms $\frac{\partial \mathbf{r}_i}{\partial \mathbf{c}_i}$ need to be calculated accordingly.
In contrast, when using the L2 loss they take a trivial form of $\frac{\partial \mathbf{r}_i}{\partial \mathbf{c}_i} {=} 1$.
In the following, we show that we can simplify the calculation of $\frac{\partial \mathbf{r}_i}{\partial \mathbf{x}_k}$ in the kernels by precomputing $\frac{\partial \mathbf{r}_i}{\partial \mathbf{c}_i}$.

The $k$-th element of $\mathbf{g}$ is calculated as:
\begin{align}
\label{eq:apply-jtj}
\mathbf{g}_k = (\mathbf{J}^T \mathbf{J} \mathbf{p})_k = \sum_{i=0}^{N} \frac{\partial \mathbf{r}_i}{\partial \mathbf{x}_k} \sum_{j=0}^{M} \frac{\partial \mathbf{r}_i}{\partial \mathbf{x}_j} \mathbf{p}_j
\end{align}
By substituting \cref{eq:res-sq-root-weights} into \cref{eq:apply-jtj}, we factor out $\frac{\partial \mathbf{r}_i}{\partial \mathbf{c}_i}$:
\begin{align}
\label{eq:apply-jtj-simplified}
\mathbf{g}_k = \sum_{i=0}^{N} (\frac{\partial \mathbf{r}_i}{\partial \mathbf{c}_i})^2 \frac{\partial \mathbf{c}_i}{\partial \mathbf{x}_k} \sum_{j=0}^{M} \frac{\partial \mathbf{c}_i}{\partial \mathbf{x}_j} \mathbf{p}_j
\end{align}
The terms $\frac{\partial \mathbf{c}_i}{\partial \mathbf{x}_k}$ are identical for a residual of the same pixel and color channel that corresponds to either the L1 or SSIM loss terms, respectively.
To avoid computing $\frac{\partial \mathbf{c}_i}{\partial \mathbf{x}_k}$ twice (and therefore doubling the grid size of all kernels), we instead sum up the contribution of both loss terms: 
\begin{align}
\label{eq:res-sq-root-weights-sum}
(\frac{\partial \mathbf{r}_i}{\partial \mathbf{c}_i})^2 = (\frac{\partial \mathbf{r}_i^1}{\partial \mathbf{c}_i})^2 + (\frac{\partial \mathbf{r}_i^2}{\partial \mathbf{c}_i})^2
\end{align}
where $\mathbf{r}_i^1$ corresponds to the $i$-th L1 residual and $\mathbf{r}_i^2$ to the $i$-th SSIM residual.
Additionally, we implement the multiplication with $(\frac{\partial \mathbf{r}_i}{\partial \mathbf{c}_i})^2$ in \cref{eq:apply-jtj-simplified} as elementwise vector product (denoted by $\odot$) of $\mathbf{\hat{u}} {=} [\sum_{j=0}^{M} \frac{\partial \mathbf{c}_0}{\partial \mathbf{x}_j} \mathbf{p}_j ... \sum_{j=0}^{M} \frac{\partial \mathbf{c}_N}{\partial \mathbf{x}_j} \mathbf{p}_j]$ and $\mathbf{\nabla r} {=} [(\frac{\partial \mathbf{r}_0}{\partial \mathbf{c}_0})^2 ... (\frac{\partial \mathbf{r}_N}{\partial \mathbf{c}_N})^2]$:
\begin{align}
\label{eq:apply-jtj-simplified-vec-vec}
\mathbf{g}_k = \sum_{i=0}^{N} \frac{\partial \mathbf{c}_i}{\partial \mathbf{x}_k} (\mathbf{\hat{u} \odot \nabla r})_i
\end{align}
This avoids additional uncoalesced global memory reads to $\mathbf{\nabla r}$ in the CUDA kernels.
Instead, we calculate $\mathbf{\hat{u} \odot \nabla r}$ in a separate operation after the \texttt{applyJ} kernel and before \texttt{applyJT}.
This also simplifies the kernels, since they now only need to compute $\frac{\partial \mathbf{c}_i}{\partial \mathbf{x}_k}$ instead of $\frac{\partial \mathbf{r}_i}{\partial \mathbf{x}_k}$.
Therefore, the only runtime overhead of using the L1 and SSIM residual terms over the L2 residuals is the computation of $\mathbf{\nabla r}$.
However, this can be efficiently computed using backpropagation (autograd) and is therefore not a bottleneck.

\section{Derivation Of Image Subsampling Weights}
We subsample batches of images to decrease the size of the gradient cache (see \cref{subsec:gsgn-kernels}).
To combine the update vectors from multiple batches, we calculate their weighted mean, as detailed in \cref{eq:combine-pcg}.
This weighted mean approximates the ``true'' solution to the normal equations (\cref{eq:norm-eq}), that does not rely on any image subsampling (and instead uses all available training images).
When subsampling images, we split the number of total residuals into smaller chunks.
In the following, we consider the case of two chunks (labeled as $_1$ and $_2$), but the same applies to any number of chunks.
We re-write the normal equations (without subsampling) using the chunk notation as:
\begin{align}
\label{eq:split-norm-eq}
    \begin{bmatrix}
    \mathbf{J}^T_{1} & \mathbf{J}^T_{2}
    \end{bmatrix}
    \begin{bmatrix}
    \mathbf{J}_{1} \\
    \mathbf{J}_{2}
    \end{bmatrix}
    \Delta = 
    \begin{bmatrix}
    \mathbf{J}^T_{1} & \mathbf{J}^T_{2}
    \end{bmatrix}
    \begin{bmatrix}
    \mathbf{F}_{1}(\mathbf{x}) \\
    \mathbf{F}_{2}(\mathbf{x})
    \end{bmatrix}
\end{align}
where we drop the additional LM regularization term for clarity and divide the Jacobian and residual vector into separate matrices/vectors according to the chunks.
The solution to the normal equations is obtained by:
\begin{align}
\label{eq:split-sol}
    \Delta = (\mathbf{J}^T_{1} \mathbf{J}_{1} + \mathbf{J}^T_{2} \mathbf{J}_{2})^{-1} (\mathbf{J}^T_{1} \mathbf{F}_{1}(\mathbf{x}) + \mathbf{J}^T_{2} \mathbf{F}_{2}(\mathbf{x}))
\end{align}
In contrast, when we subsample images, we solve the normal equations separately and obtain two solutions: 
\begin{align}
\label{eq:split-sol-subsample}
    \Delta_1 = (\mathbf{J}^T_{1} \mathbf{J}_{1})^{-1} \mathbf{J}^T_{1} \mathbf{F}_{1}(\mathbf{x})
    \\
    \Delta_2 = (\mathbf{J}^T_{2} \mathbf{J}_{2})^{-1} \mathbf{J}^T_{2} \mathbf{F}_{2}(\mathbf{x})
\end{align}
We can rewrite \cref{eq:split-sol} as a weighted mean of $\Delta_1$, $\Delta_2$:
\begin{align}
\label{eq:split-sol-re}
    \Delta = 
    K^{-1} (\mathbf{J}^T_{1} \mathbf{J}_{1}) (\mathbf{J}^T_{1} \mathbf{J}_{1})^{-1} (\mathbf{J}^T_{1} \mathbf{F}_{1}(\mathbf{x})) \\ + 
    K^{-1} (\mathbf{J}^T_{2} \mathbf{J}_{2}) (\mathbf{J}^T_{2} \mathbf{J}_{2})^{-1} (\mathbf{J}^T_{2} \mathbf{F}_{2}(\mathbf{x})) \\ =
    w_1 \Delta_1 + w_2 \Delta_2
\end{align}
where $K = (\mathbf{J}^T_{1} \mathbf{J}_{1} {+} \mathbf{J}^T_{2} \mathbf{J}_{2})$, $w_1 = K^{-1} (\mathbf{J}^T_{1} \mathbf{J}_{1})$, $w_2 = K^{-1} (\mathbf{J}^T_{2} \mathbf{J}_{2})$.
Calculating these weights requires to materialize and invert $K$, which is too costly to fit in memory.
To this end, we approximate the true weights $w_1$ and $w_2$ with $\tilde{w}_1 = \text{diag}(w_1)$ and $\tilde{w}_2 = \text{diag}(w_2)$.
This directly leads to the weighted mean that we employ in \cref{eq:combine-pcg}.

\section{Detailed Runtime Analysis}
\begin{table*}
  \centering
  \setlength\tabcolsep{2pt}
  \begin{tabular}{l | c | c | c | c}
    \toprule
    Kernel & Runtime (ms) $\downarrow$ & Compute Throughput (\%)$\uparrow$ & Memory Throughput (\%)$\uparrow$ & Register Count $\downarrow$ \\
    \midrule
    \texttt{buildCache\_p1} & 31.32 & 78.56 & 78.56 & 64 \\
    \texttt{buildCache\_p2} & 0.53 & 17.43 & 87.94 & 58 \\
    \texttt{buildCache\_p3} & 4.12 & 4.54 & 73.45 & 74 \\
    \texttt{sortCacheByGaussians} & 5.04 & 61.17 & 61.17 & 18 \\
    \texttt{diagJTJ} & 41.60 & 71.13 & 71.13 & 90 \\
    \texttt{sortX} & 4.45 & 15.15 & 60.30 & 36 \\
    \texttt{applyJ} & 10.98 & 86.32 & 86.32 & 80 \\
    \texttt{applyJT\_p1} & 3.93 & 75.79 & 75.79 & 34 \\
    \texttt{applyJT\_p2} & 0.37 & 18.83 & 89.69 & 40 \\
    \texttt{applyJT\_p3} & 3.20 & 4.75 & 78.48 & 48 \\
    \bottomrule
  \end{tabular}
  \caption{
    \textbf{Profiler analysis of CUDA kernels.}
    We provide results measured on a RTX3090 GPU for building/resorting the gradient cache and running one PCG iteration on the MipNerf360~\cite{barron2022mip} ``garden'' scene with a batch size of one image.
  }
  \label{tab:profiler-analysis}
\end{table*}

We provide additional analysis of the CUDA kernels by running the \texttt{NVIDIA Nsight Compute} profiler.
We provide results in \cref{tab:profiler-analysis} measured on a RTX3090 GPU for building/resorting the gradient cache and running one PCG iteration on the MipNerf360~\cite{barron2022mip} ``garden'' scene with a batch size of one image.
We add the suffixes \texttt{\_p1, \_p2, \_p3} to signal the three kernels that we use to implement the respective operation (see \cref{subsec:add-detail-kernels}).

Comparing the runtime of the \texttt{buildCache} and \texttt{applyJT} kernels reveals the advantage of our proposed \textit{per-pixel-per-splat} parallelization pattern.
Both compute the identical Jacobian-vector product, but the \texttt{buildCache} kernel relies on the \textit{per-pixel} parallelization pattern of the original 3DGS rasterizer~\cite{kerbl3Dgaussians}.
However, we compute the result $4.8$x faster using the gradient cache in the \texttt{applyJT} kernel.
We also note that the compute and memory throughput as well as the register count of both kernels are roughly similar.
This signals that our kernel implementation is equally efficient, i.e., there are no inherent drawbacks using our proposed GPU parallelization scheme.

\section{Results Per Scene}
\begin{table}
  \centering
  \setlength\tabcolsep{2pt}
  \begin{tabular}{l|l | rrrr}
    \toprule
        \multirow{2}{*}{Method} & \multirow{2}{*}{Scene} & \multicolumn{4}{c}{MipNeRF-360~\cite{barron2022mip}} \\
        \cmidrule(l{2pt}r{2pt}){3-6}
    & & SSIM$\uparrow$ & PSNR$\uparrow$ & LPIPS$\downarrow$ & Time (s) \\
    \midrule
    3DGS~\cite{kerbl3Dgaussians} & treehill & 0.631 & 22.44 & 0.330 & 1130 \\
    + Ours & treehill & 0.633 & 22.57 & 0.334 & \textbf{836} \\
    \midrule
    3DGS~\cite{kerbl3Dgaussians} & counter & 0.905 & 28.96 & 0.202 & 1178 \\
    + Ours & counter  & 0.904 & 28.89 & 0.206 & \textbf{927} \\
    \midrule
    3DGS~\cite{kerbl3Dgaussians} & stump & 0.769 & 26.56 & 0.217 & 1234 \\
    + Ours & stump & 0.774 & 26.67 & 0.218 & \textbf{895} \\
    \midrule
    3DGS~\cite{kerbl3Dgaussians} & bonsai & 0.939 & 31.99 & 0.206 & 1034 \\
    + Ours & bonsai & 0.938 & 31.84 & 0.208 & \textbf{794} \\
    \midrule
    3DGS~\cite{kerbl3Dgaussians} & bicycle & 0.764 & 25.20 & 0.212 & 1563 \\
    + Ours & bicycle & 0.765 & 25.30 & 0.218 & \textbf{1141} \\
    \midrule
    3DGS~\cite{kerbl3Dgaussians} & kitchen & 0.925 & 31.37 & 0.128 & 1389 \\
    + Ours & kitchen & 0.924 & 31.21 & 0.128 & \textbf{1156} \\
    \midrule
    3DGS~\cite{kerbl3Dgaussians} & flowers & 0.602 & 21.49 & 0.340 & 1132 \\
    + Ours & flowers & 0.600 & 21.52 & 0.344 & \textbf{819} \\
    \midrule
    3DGS~\cite{kerbl3Dgaussians} & room & 0.917 & 31.36 & 0.221 & 1210 \\
    + Ours & room & 0.916 & 31.10 & 0.224 & \textbf{1004} \\
    \midrule
    3DGS~\cite{kerbl3Dgaussians} & garden & 0.862 & 27.23 & 0.109 & 1573 \\
    + Ours & garden & 0.863 & 27.30 & 0.110 & \textbf{1175} \\
    \bottomrule
  \end{tabular}

  \begin{tabular}{l|l | rrrr}
    \toprule
        \multirow{2}{*}{Method} & \multirow{2}{*}{Scene} & \multicolumn{4}{c}{Deep Blending~\cite{hedman2018deep}} \\
        \cmidrule(l{2pt}r{2pt}){3-6}
    & & SSIM$\uparrow$ & PSNR$\uparrow$ & LPIPS$\downarrow$ & Time (s) \\
    \midrule
    3DGS~\cite{kerbl3Dgaussians} & playroom & 0.901 & 29.90 & 0.247 & 1085 \\
    + Ours & playroom & 0.905 & 30.24 & 0.246 & \textbf{861} \\
    \midrule
    3DGS~\cite{kerbl3Dgaussians} & drjohnson & 0.898 & 29.12 & 0.246 & 1359 \\
    + Ours & drjohnson  & 0.901 & 29.23 & 0.248 & \textbf{1040} \\
    \bottomrule
  \end{tabular}

  \begin{tabular}{l|l | rrrr}
    \toprule
        \multirow{2}{*}{Method} & \multirow{2}{*}{Scene} & \multicolumn{4}{c}{Tanks \& Temples~\cite{knapitsch2017tanks}} \\
        \cmidrule(l{2pt}r{2pt}){3-6}
    & & SSIM$\uparrow$ & PSNR$\uparrow$ & LPIPS$\downarrow$ & Time (s) \\
    \midrule
    3DGS~\cite{kerbl3Dgaussians} & train & 0.811 & 21.95 & 0.209 & 636 \\
    + Ours & train & 0.811 & 22.07 & 0.214 & \textbf{579} \\
    \midrule
    3DGS~\cite{kerbl3Dgaussians} & truck & 0.877 & 25.40 & 0.148 & 837 \\
    + Ours & truck  & 0.876 & 25.36 & 0.151 & \textbf{747} \\
    \bottomrule
  \end{tabular}
  
  \caption{
    \textbf{Quantitative comparison of our method and baselines.}
    We show the per-scene breakdown of all metrics against the 3DGS~\cite{kerbl3Dgaussians} baseline.
  }
  \label{tab:quant-comparison-all-360v2-3dgs}
\end{table}

\begin{table}
  \centering
  \setlength\tabcolsep{2pt}
  \begin{tabular}{l|l | rrrr}
    \toprule
        \multirow{2}{*}{Method} & \multirow{2}{*}{Scene} & \multicolumn{4}{c}{MipNeRF-360~\cite{barron2022mip}} \\
        \cmidrule(l{2pt}r{2pt}){3-6}
    & & SSIM$\uparrow$ & PSNR$\uparrow$ & LPIPS$\downarrow$ & Time (s) \\
    \midrule
    DISTWAR~\cite{durvasula2023distwar} & treehill & 0.633 & 22.47 & 0.327 & 898 \\
    + Ours & treehill & 0.635 & 22.54 & 0.332 & \textbf{669} \\
    \midrule
    DISTWAR~\cite{durvasula2023distwar} & counter & 0.905 & 29.00 & 0.203 & 790 \\
    + Ours & counter  & 0.904 & 28.91 & 0.205 & \textbf{687} \\
    \midrule
    DISTWAR~\cite{durvasula2023distwar} & stump & 0.771 & 26.60 & 0.216 & 1017 \\
    + Ours & stump & 0.773 & 26.70 & 0.217 & \textbf{760} \\
    \midrule
    DISTWAR~\cite{durvasula2023distwar} & bonsai & 0.939 & 32.13 & 0.206 & 677 \\
    + Ours & bonsai & 0.938 & 31.92 & 0.208 & \textbf{578} \\
    \midrule
    DISTWAR~\cite{durvasula2023distwar} & bicycle & 0.763 & 25.19 & 0.212 & 1333 \\
    + Ours & bicycle & 0.764 & 25.26 & 0.218 & \textbf{971} \\
    \midrule
    DISTWAR~\cite{durvasula2023distwar} & kitchen & 0.925 & 31.31 & 0.127 & 957 \\
    + Ours & kitchen & 0.924 & 31.14 & 0.128 & \textbf{838} \\
    \midrule
    DISTWAR~\cite{durvasula2023distwar} & flowers & 0.602 & 21.45 & 0.340 & 884 \\
    + Ours & flowers & 0.596 & 21.48 & 0.348 & \textbf{671} \\
    \midrule
    DISTWAR~\cite{durvasula2023distwar} & room & 0.916 & 31.41 & 0.221 & 803 \\
    + Ours & room & 0.916 & 31.40 & 0.224 & \textbf{680} \\
    \midrule
    DISTWAR~\cite{durvasula2023distwar} & garden & 0.862 & 27.23 & 0.109 & 1338 \\
    + Ours & garden & 0.861 & 27.32 & 0.112 & \textbf{1023} \\
    \bottomrule
  \end{tabular}

  \begin{tabular}{l|l | rrrr}
    \toprule
        \multirow{2}{*}{Method} & \multirow{2}{*}{Scene} & \multicolumn{4}{c}{Deep Blending~\cite{hedman2018deep}} \\
        \cmidrule(l{2pt}r{2pt}){3-6}
    & & SSIM$\uparrow$ & PSNR$\uparrow$ & LPIPS$\downarrow$ & Time (s) \\
    \midrule
    DISTWAR~\cite{durvasula2023distwar} & playroom & 0.900 & 29.81 & 0.247 & 729 \\
    + Ours & playroom & 0.905 & 30.24 & 0.246 & \textbf{586} \\
    \midrule
    DISTWAR~\cite{durvasula2023distwar} & drjohnson & 0.898 & 29.13 & 0.247 & 953 \\
    + Ours & drjohnson  & 0.901 & 29.13 & 0.249 & \textbf{758} \\
    \bottomrule
  \end{tabular}

  \begin{tabular}{l|l | rrrr}
    \toprule
        \multirow{2}{*}{Method} & \multirow{2}{*}{Scene} & \multicolumn{4}{c}{Tanks \& Temples~\cite{knapitsch2017tanks}} \\
        \cmidrule(l{2pt}r{2pt}){3-6}
    & & SSIM$\uparrow$ & PSNR$\uparrow$ & LPIPS$\downarrow$ & Time (s) \\
    \midrule
    DISTWAR~\cite{durvasula2023distwar} & train & 0.812 & 22.05 & 0.209 & 504 \\
    + Ours & train & 0.810 & 22.10 & 0.216 & \textbf{440} \\
    \midrule
    DISTWAR~\cite{durvasula2023distwar} & truck & 0.877 & 25.29 & 0.148 & 698 \\
    + Ours & truck  & 0.877 & 25.28 & 0.150 & \textbf{635} \\
    \bottomrule
  \end{tabular}
  
  \caption{
    \textbf{Quantitative comparison of our method and baselines.}
    We show the per-scene breakdown of all metrics against the \mbox{DISTWAR~\cite{durvasula2023distwar}} baseline.
  }
  \label{tab:quant-comparison-all-360v2-distwar}
\end{table}

\begin{table}
  \centering
  \setlength\tabcolsep{2pt}
  \begin{tabular}{l|l | rrrr}
    \toprule
        \multirow{2}{*}{Method} & \multirow{2}{*}{Scene} & \multicolumn{4}{c}{MipNeRF-360~\cite{barron2022mip}} \\
        \cmidrule(l{2pt}r{2pt}){3-6}
    & & SSIM$\uparrow$ & PSNR$\uparrow$ & LPIPS$\downarrow$ & Time (s) \\
    \midrule
    gsplat~\cite{ye2024gsplatopensourcelibrarygaussian} & treehill & 0.634 & 22.44 & 0.324 & 973 \\
    + Ours & treehill & 0.635 & 22.54 & 0.332 & \textbf{701} \\
    \midrule
    gsplat~\cite{ye2024gsplatopensourcelibrarygaussian} & counter & 0.908 & 28.99 & 0.201 & 903 \\
    + Ours & counter  & 0.904 & 28.91 & 0.205 & \textbf{762} \\
    \midrule
    gsplat~\cite{ye2024gsplatopensourcelibrarygaussian} & stump & 0.769 & 26.53 & 0.218 & 1097 \\
    + Ours & stump & 0.774 & 26.70 & 0.217 & \textbf{793} \\
    \midrule
    gsplat~\cite{ye2024gsplatopensourcelibrarygaussian} & bonsai & 0.937 & 31.95 & 0.208 & 783 \\
    + Ours & bonsai & 0.938 & 31.92 & 0.208 & \textbf{646} \\
    \midrule
    gsplat~\cite{ye2024gsplatopensourcelibrarygaussian} & bicycle & 0.765 & 25.21 & 0.206 & 1398 \\
    + Ours & bicycle & 0.765 & 25.26 & 0.218 & \textbf{988} \\
    \midrule
    gsplat~\cite{ye2024gsplatopensourcelibrarygaussian} & kitchen & 0.926 & 31.17 & 0.128 & 1086 \\
    + Ours & kitchen & 0.924 & 31.14 & 0.128 & \textbf{921} \\
    \midrule
    gsplat~\cite{ye2024gsplatopensourcelibrarygaussian} & flowers & 0.600 & 21.53 & 0.338 & 965 \\
    + Ours & flowers & 0.601 & 21.48 & 0.348 & \textbf{709} \\
    \midrule
    gsplat~\cite{ye2024gsplatopensourcelibrarygaussian} & room & 0.920 & 31.48 & 0.219 & 913 \\
    + Ours & room & 0.916 & 31.39 & 0.224 & \textbf{753} \\
    \midrule
    gsplat~\cite{ye2024gsplatopensourcelibrarygaussian} & garden & 0.869 & 27.48 & 0.105 & 1462 \\
    + Ours & garden & 0.861 & 27.32 & 0.112 & \textbf{1085} \\
    \bottomrule
  \end{tabular}

  \begin{tabular}{l|l | rrrr}
    \toprule
        \multirow{2}{*}{Method} & \multirow{2}{*}{Scene} & \multicolumn{4}{c}{Deep Blending~\cite{hedman2018deep}} \\
        \cmidrule(l{2pt}r{2pt}){3-6}
    & & SSIM$\uparrow$ & PSNR$\uparrow$ & LPIPS$\downarrow$ & Time (s) \\
    \midrule
    gsplat~\cite{ye2024gsplatopensourcelibrarygaussian} & playroom & 0.907 & 29.89 & 0.248 & 799 \\
    + Ours & playroom & 0.904 & 30.90 & 0.247 & \textbf{626} \\
    \midrule
    gsplat~\cite{ye2024gsplatopensourcelibrarygaussian} & drjohnson & 0.901 & 29.16 & 0.244 & 1040 \\
    + Ours & drjohnson  & 0.901 & 29.07 & 0.251 & \textbf{805} \\
    \bottomrule
  \end{tabular}

  \begin{tabular}{l|l | rrrr}
    \toprule
        \multirow{2}{*}{Method} & \multirow{2}{*}{Scene} & \multicolumn{4}{c}{Tanks \& Temples~\cite{knapitsch2017tanks}} \\
        \cmidrule(l{2pt}r{2pt}){3-6}
    & & SSIM$\uparrow$ & PSNR$\uparrow$ & LPIPS$\downarrow$ & Time (s) \\
    \midrule
    gsplat~\cite{ye2024gsplatopensourcelibrarygaussian} & train & 0.811 & 21.64 & 0.209 & 558 \\
    + Ours & train & 0.809 & 22.09 & 0.216 & \textbf{381} \\
    \midrule
    gsplat~\cite{ye2024gsplatopensourcelibrarygaussian} & truck & 0.880 & 25.35 & 0.149 & 735 \\
    + Ours & truck  & 0.877 & 25.28 & 0.150 & \textbf{447} \\
    \bottomrule
  \end{tabular}
  \caption{
    \textbf{Quantitative comparison of our method and baselines.}
    We show the per-scene breakdown of all metrics against the gsplat~\cite{ye2024gsplatopensourcelibrarygaussian} baseline.
  }
  \label{tab:quant-comparison-all-360v2-gsplat}
\end{table}

\begin{table}
  \centering
  \setlength\tabcolsep{2pt}
  \begin{tabular}{l|l | rrrrr}
    \toprule
        \multirow{2}{*}{Method} & \multirow{2}{*}{Scene} & \multicolumn{4}{c}{MipNeRF-360~\cite{barron2022mip}} \\
        \cmidrule(l{2pt}r{2pt}){3-7}
    & & SSIM$\uparrow$ & PSNR$\uparrow$ & LPIPS$\downarrow$ & Time (s) & \#G (M) \\
    \midrule
    Taming~\cite{mallick2024taming} & treehill & 0.625 & 23.00 & 0.385 & 479 & 0.78\\
    + Ours & treehill & 0.623 & 23.03 & 0.332 & \textbf{381} & 0.78 \\
    \midrule
    Taming~\cite{mallick2024taming} & counter & 0.896 & 28.59 & 0.223 & 646 & 0.31 \\
    + Ours & counter  & 0.894 & 28.51 & 0.205 & \textbf{537} & 0.31 \\
    \midrule
    Taming~\cite{mallick2024taming} & stump & 0.734 & 25.96 & 0.292 & 405 & 0.48 \\
    + Ours & stump & 0.733 & 25.98 & 0.217 & \textbf{315} & 0.48 \\
    \midrule
    Taming~\cite{mallick2024taming} & bonsai & 0.934 & 31.73 & 0.221 & 634 & 0.41 \\
    + Ours & bonsai & 0.932 & 31.64 & 0.208 & \textbf{504} & 0.41 \\
    \midrule
    Taming~\cite{mallick2024taming} & bicycle & 0.716 & 24.78 & 0.295 & 485 & 0.81 \\
    + Ours & bicycle & 0.709 & 24.75 & 0.218 & \textbf{376} & 0.81 \\
    \midrule
    Taming~\cite{mallick2024taming} & kitchen & 0.918 & 30.85 & 0.141 & 722 & 0.48 \\
    + Ours & kitchen & 0.918 & 30.84 & 0.128 & \textbf{597} & 0.48 \\
    \midrule
    Taming~\cite{mallick2024taming} & flowers & 0.554 & 21.00 & 0.407 & 465 & 0.57\\
    + Ours & flowers & 0.549 & 20.99 & 0.348 & \textbf{365} & 0.57 \\
    \midrule
    Taming~\cite{mallick2024taming} & room & 0.906 & 31.12 & 0.251 & 621 & 0.22 \\
    + Ours & room & 0.906 & 31.21 & 0.224 & \textbf{521} & 0.22 \\
    \midrule
    Taming~\cite{mallick2024taming} & garden & 0.852 & 27.24 & 0.128 & 638 & 1.90 \\
    + Ours & garden & 0.852 & 27.25 & 0.112 & \textbf{483} & 1.90 \\
    \bottomrule
  \end{tabular}

  \begin{tabular}{l|l | rrrrr}
    \toprule
        \multirow{2}{*}{Method} & \multirow{2}{*}{Scene} & \multicolumn{4}{c}{Deep Blending~\cite{hedman2018deep}} \\
        \cmidrule(l{2pt}r{2pt}){3-7}
    & & SSIM$\uparrow$ & PSNR$\uparrow$ & LPIPS$\downarrow$ & Time (s) & \#G (M) \\
    \midrule
    Taming~\cite{mallick2024taming} & playroom & 0.900 & 30.29 & 0.278 & 419 & 0.18 \\
    + Ours & playroom & 0.902 & 30.41 & 0.280 & \textbf{324} & 0.18 \\
    \midrule
    Taming~\cite{mallick2024taming} & drjohnson & 0.899 & 29.38 & 0.269 & 475 & 0.40 \\
    + Ours & drjohnson  & 0.899 & 29.40 & 0.271 & \textbf{370} & 0.40 \\
    \bottomrule
  \end{tabular}

  \begin{tabular}{l|l | rrrrr}
    \toprule
        \multirow{2}{*}{Method} & \multirow{2}{*}{Scene} & \multicolumn{4}{c}{Tanks \& Temples~\cite{knapitsch2017tanks}} \\
        \cmidrule(l{2pt}r{2pt}){3-7}
    & & SSIM$\uparrow$ & PSNR$\uparrow$ & LPIPS$\downarrow$ & Time (s) & \#G (M) \\
    \midrule
    Taming~\cite{mallick2024taming} & train & 0.815 & 22.18 & 0.205 & 411 & 0.36 \\
    + Ours & train & 0.802 & 22.28 & 0.241 & \textbf{328} & 0.36 \\
    \midrule
    Taming~\cite{mallick2024taming} & truck & 0.879 & 25.40 & 0.146 & 514 & 0.27 \\
    + Ours & truck  & 0.862 & 25.16 & 0.178 & \textbf{292} & 0.27 \\
    \bottomrule
  \end{tabular}
  \caption{
    \textbf{Quantitative comparison of our method and baselines.}
    We show the per-scene breakdown of all metrics against the Taming-3DGS~\cite{mallick2024taming} baseline.
    Additionally, we include the number of Gaussians in millions ($\text{\#G (M)}$) that we obtained using the default hyperparameters for ``budgeting''.
  }
  \label{tab:quant-comparison-all-360v2-taming-3dgs}
\end{table}

We provide a per-scene breakdown of our main quantitative results against all baselines on all datasets.
The comparisons against 3DGS~\cite{kerbl3Dgaussians} are in \cref{tab:quant-comparison-all-360v2-3dgs}.
The comparisons against DISTWAR~\cite{durvasula2023distwar} are in \cref{tab:quant-comparison-all-360v2-distwar}.
The comparisons against gsplat~\cite{ye2024gsplatopensourcelibrarygaussian} are in \cref{tab:quant-comparison-all-360v2-gsplat}.
The comparisons against Taming-3DGS~\cite{mallick2024taming} are in \cref{tab:quant-comparison-all-360v2-taming-3dgs}.
Our method shows consistent acceleration of the optimization runtime on all scenes, while achieving the same quality.

\end{document}